%% file: main.tex
\PassOptionsToPackage{dvipsnames,table}{xcolor}
\documentclass[]{pediamed-ai}

\title{Decoding Children's Gait Behavior}

\author[1,2,*]{Yifan Shen}
\author[1,*]{Boyi Li}
\author[2,3,4,*]{Meihuan Huang}
\author[1,*]{Yuanzhe Liu}
\author[1,2,*,\S]{Xu Cao}
\author[1]{Jinyang Jin}
\author[1]{Zhengyuan Li}
\author[5]{Anglin Liu}
\author[1]{Junho Kim}
\author[2]{Jingyuan Zhu}
\author[2]{Fangzhou Lan}
\author[2,3]{Jianguo Cao}
\author[5]{Jintai Chen}
\author[1]{Ismini Lourentzou}
\author[1,\dagger]{James M. Rehg}

\affiliation[1]{University of Illinois Urbana-Champaign}
\affiliation[2]{PediaMed AI}
\affiliation[3]{Shenzhen Children's Hospital}
\affiliation[4]{Hong Kong Polytechnic University}
\affiliation[5]{The Hong Kong University of Science and Technology (Guangzhou)}

\contribution[*]{Equal contribution}
\contribution[\S]{Project lead}
\contribution[\dagger]{Corresponding author}

\input{text_boxes/inc_macros}

\input{math_commands.tex}

\usepackage{array}
\usepackage{color}
\usepackage{epsfig}
\usepackage{graphicx}

\usepackage{booktabs}
\usepackage{tabularx}
\usepackage{ltablex}
\usepackage{tabularray}
\newcolumntype{C}{>{\centering\arraybackslash}X}
\usepackage{multirow}
\usepackage{diagbox}
\usepackage{hhline}

\usepackage{extarrows}
\usepackage{makecell}
\usepackage{wrapfig}
\usepackage{colortbl}
\usepackage{longtable}
\usepackage{fancyvrb}
\usepackage{listings}

\usepackage{adjustbox}
\usepackage{array}

\usepackage{amsmath,amsfonts,amssymb}
\usepackage{bm}
\usepackage{nicefrac}
\usepackage{microtype}

\usepackage{changepage}
\usepackage{extramarks}
\usepackage{fancyhdr}
\usepackage{setspace}
\usepackage{soul}
\usepackage{xspace}
\usepackage{multicol}

\usepackage{url}

\usepackage{enumerate}
\usepackage{enumitem}
\setlist[itemize]{leftmargin=*}

\usepackage{pifont}

\usepackage{longtable}
\usepackage{ragged2e}
\usepackage[most]{tcolorbox}

\usepackage{algorithm,algpseudocode}

\usepackage[symbol]{footmisc}

\usepackage[most]{tcolorbox}

\usepackage{caption}
\usepackage{scalefnt}
\usepackage{fontawesome5}

\usepackage{titletoc}
\definecolor{citecolor}{HTML}{0071BC}
\definecolor{linkcolor}{HTML}{ED1C24}
\usepackage{hyperref}

\usepackage{marginnote}

\usepackage{lipsum}
\usepackage{nicematrix}
\usepackage{subcaption}

\usepackage{siunitx}
\usepackage{calc}
\usepackage{cleveref}
\crefname{appendix}{Appendix}{Appendices}
\crefname{section}{Section}{Sections}
\crefname{figure}{Fig.}{Figs.}
\crefname{table}{Tab.}{Tabs.}

\usepackage{titlecaps}
\usepackage{wasysym}

\usepackage{etoolbox}

\newcommand{\eg}{e.g.}

\definecolor{ChildC}{HTML}{2F6DB3}   % blue
\definecolor{ChildA}{HTML}{6A4FB3}   % purple
\definecolor{ChildE}{HTML}{C05A9D}   % magenta

\newcommand{\kindatiny}{\fontsize{6pt}{7.2pt}\selectfont}
\newlength\savewidth

\definecolor{qcolor}{HTML}{536872}

\newcolumntype{P}[1]{>{\centering\arraybackslash}p{#1}}

\newcommand{\tablestyle}[2]{%
	\fontfamily{ptm}\selectfont%
	\let\itold\it%
	\def\it{\itold \fontfamily{ptm}\selectfont}%
	\setlength{\tabcolsep}{#1}\renewcommand{\arraystretch}{#2}\centering\kindatiny%
	\let\citeold\cite%
	\renewcommand{\cite}[1]{\normalfont\fontfamily{ptm}\selectfont\tiny\citeold{##1}}%
}
\newcolumntype{Y}{>{\centering\arraybackslash}X}

\titlecontents{section}
[1.5em]
{\addvspace{-0.5pt}}
{\bfseries\contentslabel{2.3em}}
{\hspace*{-2.3em}\bfseries}
{\bfseries\titlerule*[.5pc]{.}\contentspage}
\titlecontents{subsection}
[3.8em]
{\addvspace{-2.2pt}}
{\contentslabel{2.3em}}
{\hspace*{-2.3em}}
{\titlerule*[.5pc]{.}\contentspage}

\newtcolorbox{planbox}[1]{
    colback=gray!5,
    colframe=gray!75,
    title=#1,
    fonttitle=\bfseries
}

\tcbset{colback=white}
\colorlet{titleblue}{blue!80!black}
\colorlet{titlered}{red!80!black}
\colorlet{titlegreen}{green!80!black}
\colorlet{darkgreen}{green!50!black}

\definecolor{logoRed}{HTML}{CB2F10}
\definecolor{logoBlue}{HTML}{1A4E8A}
\definecolor{logoCyan}{HTML}{56BBCC}

\definecolor{genLLM}{RGB}{250,250,250}
\definecolor{specLLM}{RGB}{230,230,230}
\definecolor{ourLLM}{RGB}{230,242,230}
\definecolor{spaLLM}{RGB}{255,249,196}

\definecolor{genVLM}{RGB}{255,255,255}
\definecolor{ourVLM}{RGB}{210,210,210}
\DeclareCaptionFont{times}{\fontfamily{ptm}\selectfont}

\tcbset{
  aibox/.style={
    width=\linewidth,
    top=10pt,
    colback=white,
    colframe=black,
    colbacktitle=black,
    enhanced,
    center,
    attach boxed title to top left={yshift=-0.1in,xshift=0.15in},
    boxed title style={boxrule=0pt,colframe=white,},
  }
}

\newtcolorbox{AIbox}[2][]{aibox,title=#2,#1}

\usepackage{tikz}
\makeatletter
\definecolor{applegreen}{rgb}{0.55, 0.71, 0.0}

\newcommand{\mrot}[1]{\rlap{\rotatebox[origin=lb]{45}{#1}}}

\abstract{
We introduce a new problem domain for human action recognition: the fine-grained analysis of children's gait behaviors from standard RGB video. We specifically target the ambulatory patterns of children aged 3-17 years. Such behaviors arise naturally in the diagnosis and treatment of several critical developmental and neuromuscular disorders, such as cerebral palsy and hemiplegia. Despite their clinical value, current 3D sensor-based gait analysis systems are expensive, intrusive, and often impractical for young subjects. To address this, we introduce a new dataset comprising over 1,100 high-frame-rate (60 FPS) video sequences from 110 subjects, accompanied by synchronized, anonymized pose sequences. In each session, the child performs a 5-second "walk-around" task, capturing the gait cycle from multiple viewpoints. Crucially, we demonstrate that current state-of-the-art approaches, including gait foundation models and Multimodal Large Language Models (MLLMs), fail to effectively resolve these clinical nuances. We identify the key technical challenges in analyzing these erratic and subtle motor patterns and describe a unified end-to-end framework for decoding fundamental components of pediatric gait. Through comprehensive experimental results, we demonstrate the potential of this dataset to drive novel research questions and establish a rigorous baseline for automated child gait assessment. 
}

\metadata[Website]{\url{https://pediamedai.com/ChildrenGait/}}
\metadata[HuggingFace]{\url{https://huggingface.co/datasets/PediaMedAI/ChildrenGait}}
\metadata[Correspondence to]{xucao@pediamed.ai, jrehg@illinois.edu}

\definecolor{lightgray}{rgb}{0.95, 0.95, 0.95}

\definecolor{baselinecolor}{gray}{.9}

\begin{document}

\maketitle

\input{sec/1_intro}
\input{sec/2_related_works}
\input{sec/3_dataset}
\input{sec/5_decoding}
\input{sec/7_conclusion}

\clearpage
\bibliography{main}
\bibliographystyle{bibstyle}

\newpage
\input{sec/X_suppl}

\end{document}

%% file: text_boxes/inc_macros.tex
\newcolumntype{L}[1]{>{\raggedright\let\newline\\\arraybackslash\hspace{0pt}}m{#1}}
\newcolumntype{R}[1]{>{\raggedleft\let\newline\\\arraybackslash\hspace{0pt}}m{#1}}

\newcommand{\ignore}[1]{}

\makeatletter
\DeclareRobustCommand\onedot{\futurelet\@let@token\@onedot}
\def\@onedot{\ifx\@let@token.\else.\null\fi\xspace}

\makeatother

\definecolor{MyBlue}{rgb}{0.46, 0.50, 0.61}
\definecolor{MyDarkBlue}{rgb}{0,0.08,0.8}
\definecolor{MyDarkGreen}{RGB}{45,155,45}
\definecolor{MyDarkRed}{rgb}{0.8,0.02,0.02}
\definecolor{MyOrange}{rgb}{1.0, 0.4, 0.2}
\definecolor{MyPurple}{RGB}{111,0,255}
\definecolor{MyRed}{rgb}{0.8,0.0,0.0}
\definecolor{MyGold}{rgb}{0.75,0.6,0.12}
\definecolor{MyDarkgray}{rgb}{0.66, 0.66, 0.66}
\definecolor{MyBrown}{rgb}{0.65, 0.16, 0.16}
\definecolor{MyMutedRose}{rgb}{0.58, 0.29, 0.35}
\definecolor{JiayuanColor}{rgb}{0.60,0.43,0.48}
\definecolor{erranColor}{rgb}{24, 40, 113}

\definecolor{citecolor}{HTML}{696FAD}

\definecolor{bggray}{HTML}{F5F5F5}
\definecolor{pvdblue}{HTML}{DAE8FC}
\definecolor{RoseQuartzBg}{HTML}{F7CAC9}
\definecolor{RoseQuartz}{HTML}{F5A798}
\definecolor{Serenity}{HTML}{92A8D1}
\definecolor{OrangeRed}{rgb}{1.0, 0.27, 0.0}
\definecolor{RoyalBlue}{cmyk}{1, 0.50, 0, 0}
\definecolor{Turquoise}{HTML}{0F4C81}
\definecolor{mint}{rgb}{0.24, 0.71, 0.54}
\definecolor{green}{rgb}{0.0, 0.120, 0.0}

\newdimen\abovecrulesep
\newdimen\belowcrulesep
\makeatletter
\patchcmd{\@@@cmidrule}{\aboverulesep}{\abovecrulesep}{}{}
\patchcmd{\@xcmidrule}{\belowrulesep}{\belowcrulesep}{}{}
\makeatother

\definecolor{mybluetitle}{HTML}{4B527E} %

\definecolor{codegreen}{HTML}{478058}%
\definecolor{codegray}{rgb}{0.5,0.5,0.5}
\definecolor{codepurple}{HTML}{4F5E80} %
\definecolor{backcolour}{rgb}{0.95,0.95,0.92}
\lstdefinestyle{mystyle}{
    backgroundcolor=\color{backcolour},
    commentstyle=\color{codegreen},
    keywordstyle=\color{magenta},
    numberstyle=\tiny\color{codegray},
    stringstyle=\color{codepurple},
    basicstyle=\ttfamily\scriptsize,
    breakatwhitespace=false,
    breaklines=true,
    captionpos=b,
    keepspaces=true,
    frame=none,
    numbersep=5pt,
    showspaces=false,
    showstringspaces=false,
    showtabs=false,
    tabsize=2
}

\newtcolorbox{promptbox}[2][]{
    enhanced, 
    breakable,
    center title,
    left*=0pt, right*=0pt,
    boxsep=2pt, left=5pt, right=5pt,
    skin first=enhanced,
    skin middle=enhanced,
    skin last=enhanced,
    colback  = backcolour,
    fonttitle=\bfseries\rmfamily,
    fontupper=\scriptsize,
    title={\footnotesize\strut{#2}},
    #1
    }

\newtcolorbox{onebox}[2][]{
    enhanced, 
    center title,
    left*=0pt, right*=0pt,
    boxsep=2pt, left=5pt, right=5pt,
    skin first=enhanced,
    skin middle=enhanced,
    skin last=enhanced,
    colframe = mybluetitle!90,
  colback  = mybluetitle!10,
    fonttitle=\bfseries\rmfamily\fontfamily{phv}\selectfont,
    title={\footnotesize\strut{#2}  \refstepcounter{subsubsection} \addcontentsline{toc}{subsubsection}{\string\numberline{\thesubsubsection}#2}
    },
    #1
    }

%% file: math_commands.tex
\usepackage{amsmath,amsfonts,bm}

\def\eqref#1{equation~\ref{#1}}

\def\1{\bm{1}}

\DeclareMathAlphabet{\mathsfit}{\encodingdefault}{\sfdefault}{m}{sl}
\SetMathAlphabet{\mathsfit}{bold}{\encodingdefault}{\sfdefault}{bx}{n}

%% file: sec/1_intro.tex
\section{Introduction}
\label{sec:intro}

Quantitative gait analysis constitutes a fundamental pillar of clinical diagnostics and physical rehabilitation for movement disorders~\cite{tan2026gaitdynamics, baker2016gait, bonanno2023gait, sharma2024factors, ben2023quantitative}. In pediatric populations, the challenge of decoding pediatric gait across different developmental milestones is central to the diagnosis and management of a variety of developmental disorders~\cite{armand2016gait,rathinam2014observational,baker2006gait}. Moreover, with the advent of  Brain-Computer Interfaces (BCI) and neuroprosthetics, objective gait quantification has evolved into a critical benchmark for evaluating the efficacy of functional restoration in child patients~\cite{do2013brain,blanco2024gait,shen2026survey}. Currently, the early identification of gait abnormalities in conditions such as cerebral palsy (CP) relies heavily on subjective visual assessment by a pediatrician during a standard office visit~\cite{maathuis2005gait,dickens2006validation}. 
While research utilizing 3D gait analysis and Inertial Measurement Units (IMUs) has successfully identified quantitative warning signs for gait deviations in cerebral palsy~\cite{cook2003gait,deluca1997alterations}, deploying these technologies in clinical practice faces significant hurdles. Technologies that require significant physical space (\eg, to obtain multiple unobstructed lines-of-sight in motion capture) are difficult to incorporate and can bias the adoption of technology to large, well-resourced clinical sites, creating potential inequities~\cite{states2021instrumented,lam2023systematic}. Technologies that require an attachment to children's bodies (\eg, IMU-based measurement) will not be tolerated by all children and may induce reactivity, where the measurement technology alters the natural gait behavior~\cite{bourgeois2014spatio,kanko2021concurrent}. In this setting, there is immense potential for computer vision-based movement analysis to bridge this gap, as camera hardware is relatively inexpensive and accurate measurements can, in principle, be obtained from only one or two camera views, without encumbering the child. Automated analyses could potentially scale early screening efforts by bringing reliable, rich, and non-intrusive measurement of child gait into everyday clinical settings~\cite{colyer2018review,wishaupt2024applicability}.

Despite this clinical imperative, existing computer vision methods for gait analysis have largely focused on healthy adult subjects~\cite{ranjan2025computer, chen2022computer}, relying on the assumption that gait is a mature, stable, and highly periodic process~\cite{fan2023opengait}. Such `adult-centric' foundation models fail to capture the high entropy, high intra-class variance, and inconsistent motion patterns inherent to the developing motor system~\cite{sutherland1997development,vielemeyer2026full,shen2026position}. In this work, we define a new challenge for the computer vision community: \emph{the fine-grained analysis of children's gait from video, targeting clinically-relevant characterizations of children's gait quality.}

To address this challenge, we introduce the \textbf{Children Gait Video (CGV) dataset}, a repository comprising over 1,100 video sessions from 110 child subjects. Video collection followed a standardized observational protocol: brief (3–5 second) video sequences capturing anterior, posterior, and lateral walking views of children aged 3–17 years. Videos were annotated by expert clinicians for the Edinburgh Visual Gait Score (EVGS) \cite{ong2008reliability,beynon2010correlations}. In addition to clinical assessments, CGV also contains rich, frame-level annotations, including instance segmentation masks, bounding boxes, and anatomical keypoints, all tailored to the nuances of the developing body. \looseness=-1
We also present a comprehensive evaluation of the effectiveness of modern VLMs in analyzing children's videos and inferring the EVGS, and we find that state-of-the-art models lack sensitivity to the subtle details of children's gait. To address this limitation, we introduce a new method, ChildGait-Video, and show that it achieves SoTA performance across all baselines, reaching a highest accuracy range of 70\% to 93\% in all 34 scoring items. 

\noindent In summary, this paper makes the following contributions:
\begin{itemize}[itemsep=0.5ex, parsep=0pt, topsep=0pt]
\renewcommand{\labelitemi}{$\bullet$}
    \item We introduce the Children Gait Video (CGV) dataset, which is the largest repository of multi-camera gait videos of children with developmental motor conditions, containing both clinically relevant assessments (EVGS) and rich frame-level pose annotations.
    \item We provide the first comprehensive assessment of the ability of open and closed VLMs to decode the subtle signs of gait abnormalities from video, and find that out-of-the-box models are ineffective for this task.
    \item We present ChildGait-Video, a novel video analysis method that is the current SoTA for automated inference of EVGS scores from children's videos.
\end{itemize}

%% file: sec/2_related_works.tex
\section{Related Work}

\noindent \textbf{Human Gait Datasets.}
The landscape of gait datasets has progressively evolved from general domain biometric recognition to specialized health and clinical applications~\cite{zhao2023effective,sankhla2022human}. In the general domain, foundational benchmarks such as the CASIA series and the extensive OU-ISIR database families have driven immense progress in appearance-invariant and multi-view identity recognition~\cite{xu2017isir,makihara2012isir,iwama2012isir,aman2024performance,takemura2018multi}. Other datasets, including SUSTech1K and CMU MoBo, further expand visual gait analysis into wild and treadmill environments~\cite{shen2023lidargait,yang2013face}. In addition, recent large-scale benchmarks curated for practical in-the-wild and cross-covariate evaluation have become increasingly central to gait recognition research
%, including OUMVLP, GREW, Gait3D, CCPG, GaitLU-1M
\cite{li2022multi,zhu2021gait,zheng2022gait,li2023depth,fan2023learning}.
Concurrently, the focus has shifted toward clinical and health-oriented gait analysis, evidenced by the emergence of multimodal and sensor-based datasets like WearGait-PD~\cite{anderson2026weargait}, WhuGait~\cite{zou2020deep}, and MAREA~\cite{khandelwal2017evaluation}, which target specific pathologies such as Parkinson's disease and provide quantitative biomarkers for fall risk and symptom monitoring~\cite{di2020gait}. However, despite the existence of large-scale datasets spanning wide age ranges, for example, OULP-Age \cite{xu2017isir}, there is an absence of comprehensive, multi-view video datasets tailored for clinical gait screening in children. Pediatric gait exhibits unique biomechanical developmental trajectories and specific pathological manifestations~\cite{pistacchi2017gait}, rendering the lack of dedicated pediatric multi-view datasets a critical bottleneck. 
To bridge this gap, we introduce the CGV Dataset, a new resource designed to support fine-grained analysis of children's gait and enable the development of clinically meaningful assessment models. \looseness=-1

\noindent \textbf{Gait Vision Modeling.}
Visual gait modeling has evolved significantly through deep representations. Open-source benchmarks like OpenGait~\cite{fan2023opengait,fan2025opengait} have unified classic silhouette-based architectures~\cite{chao2019gaitset,chao2021gaitset,fan2020gaitpart,hou2020gait}, while concurrent studies tackle cross-covariate and in-the-wild challenges~\cite{zou2024cross,li2023depth,zhu2021gait}. Concurrently, pose-based and skeleton-based models offer robustness against appearance changes by pairing 2D keypoint detectors~\cite{cao2019openpose,fang2022alphapose,sun2019deep,jiang2023rtmpose} with 3D pose lifting or triangulation~\cite{pavllo20193d,zheng20213d,zhu2023motionbert,zhang2022mixste,karashchuk2021anipose}. Recent specialized skeleton maps further improve baselines under viewpoint variations~\cite{fan2024skeletongait,fan2025opengait,fu2023gpgait,fu2024cut}. To model these dynamic sequences, diverse spatiotemporal encoders are utilized, including graph convolutional networks~\cite{yan2018spatial,shi2019two,liu2020disentangling,duan2022revisiting,teepe2021gaitgraph} and video-based architectures~\cite{feichtenhofer2019slowfast,tong2022videomae,shen2026egoforge,zhang2026linkedout,zhang2026thinkjepa}, alongside recent extensions into LiDAR and point-cloud modalities~\cite{shen2023lidargait}. Most recently, the field has gravitated toward Transformers~\cite{zhang2023spatial,catruna2024gaitpt} and massive foundation models~\cite{ye2025silhouette}, leveraging diffusion and large vision pipelines to achieve zero-shot generalization under real-world degradations~\cite{ye2024biggait,jin2025denoising}. However, state-of-the-art architectures are almost exclusively pre-trained on adult datasets~\cite{hulzinga2020new}. Due to substantial differences in children's skeletal proportions, applying these pre-trained models to pediatric gait introduces severe domain shifts, necessitating new, domain-specific modeling strategies.

\noindent \textbf{Children's Gait Screening and Assessment.}
Gait screening and assessment translate raw motion signals into actionable clinical diagnostics and developmental metrics. Quantitative Gait Analysis typically extracts fundamental spatiotemporal parameters, such as velocity, cadence, step length, and double support time, alongside kinetic ground reaction forces~\cite{wang2022multi,ye2025silhouette,li20261st}. While generalized functional scales like the Functional Gait Assessment (FGA)~\cite{wrisley2004reliability}, and UPDRS~\cite{martinez2013expanded} are widely used for adult fall risk and neurological evaluation, pediatric gait assessment demands specialized criteria. Observational scales are critical for diagnosing neurodevelopmental disorders such as Cerebral Palsy (CP)~\cite{armand2016gait}. Beyond the widely recognized Edinburgh Visual Gait Score (EVGS)~\cite{read2003edinburgh}, the clinical assessment includes the Visual Gait Assessment Scale (VGAS)~\cite{lord1998visual}, the Gait Deviation Index (GDI)~\cite{schwartz2008gait}, and various musculoskeletal rubrics~\cite{field2001spinal}. Modern automated screening endeavors to map video-derived 3D motion directly to these structured scores using ordinal learning techniques~\cite{cao2020rank,shi2023deep} and multi-task frameworks~\cite{luvizon20182d,luvizon2020multi,shen2026evaluating}. However, most existing recognition-oriented frameworks focus on coarse classification rather than the item-level, phase-specific observational scoring required for reliable pediatric screening, highlighting the necessity for both specialized datasets and dedicated pediatric modeling techniques. \looseness=-1

%% file: sec/3_dataset.tex
\section{Challenges in Pediatric Gait Analysis}
\label{sec:challenge}

\begin{figure}[!t]
    \centering
    \includegraphics[width=0.99\linewidth]{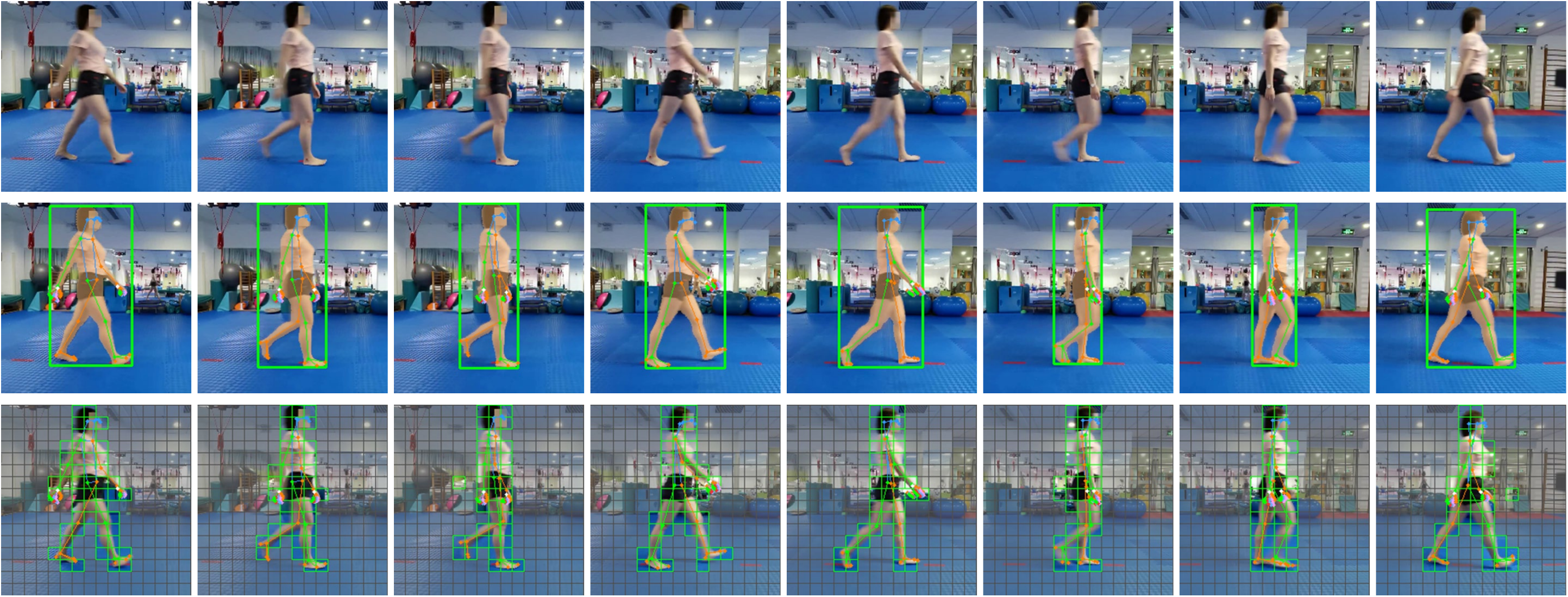}
    \caption{\textbf{Video Examples and Annotations.} \textit{Top:} An example of a raw video of a single gait cycle. \textit{Middle:} We use SAM 3 to perform instance segmentation and Sapiens-2B to perform pose estimation to obtain bounding boxes, masks, and keypoints. Video then incorporates keypoints as token-level prompts. \textit{Bottom:} We use these annotations to conduct mask-guided pruning to ignore irrelevant background noise.}
    \label{fig:visualization}
\end{figure}

From the perspective of action and activity recognition, the automated 2D video analysis of pediatric gait introduces unique technical challenges that are largely absent in existing adult-centric datasets. First, gait modeling of children confronts a significant domain shift in skeletal structure.  Significant anthropometric differences exist between pediatric and adult populations. The relative limb proportions and center of mass in children do not scale linearly from adult data. Consequently, foundation models pre-trained on adult kinematics often fail to generalize, a limitation frequently observed in 3D human mesh recovery and child pose estimation tasks~\cite{chatzichristodoulou2025age,qiu2025lhm,cao2022aggpose,hui2025infantnet}. Second, clinical gait assessment is inherently a fine-grained, phase-dependent task. Unlike the coarse disorder classification labels found in healthcare datasets like Scoliosis1K~\cite{zhou2024gait,zhou2025pose}, effective gait analysis requires the precise estimation of joint angles at specific instances of the gait cycle (\eg, maximum knee flexion during swing). Third, the capture environment involves severe occlusion and multi-person interaction. Young children frequently require physical guidance or encouragement from parents and clinicians during the walking process, resulting in complex visual clutter and frequent inter-person occlusions that challenge standard tracking and segmentation gait modeling pipelines.

Analyzing pediatric gait in therapeutic and diagnostic contexts creates a critical intersection for computer vision researchers and clinical practitioners to investigate fundamental aspects of early motor development. This collaboration opens avenues to explore profound clinical questions, such as whether subtle gait deviations can function as early indicators of autism spectrum disorder, how effectively BCI assistive technologies enhance gait kinematics, or how subsequent walking abnormalities link back to atypical infant general movements. Resolving these child-centered inquiries through a data-driven lens requires the computer vision community to pioneer new, robust methodologies capable of modeling complex, dynamic, developmental gait behaviors from unconstrained video and other sensing modalities. \looseness=-1

\section{Children Gait Video (CGV) Dataset}
\label{sec:dataset}

\input{tabs/items}

\begin{figure}[!t]
    \centering
    \begin{minipage}[c]{0.25\textwidth}
        \centering
        \begin{minipage}[c]{0.82\textwidth}
            \centering
            \includegraphics[width=\textwidth]{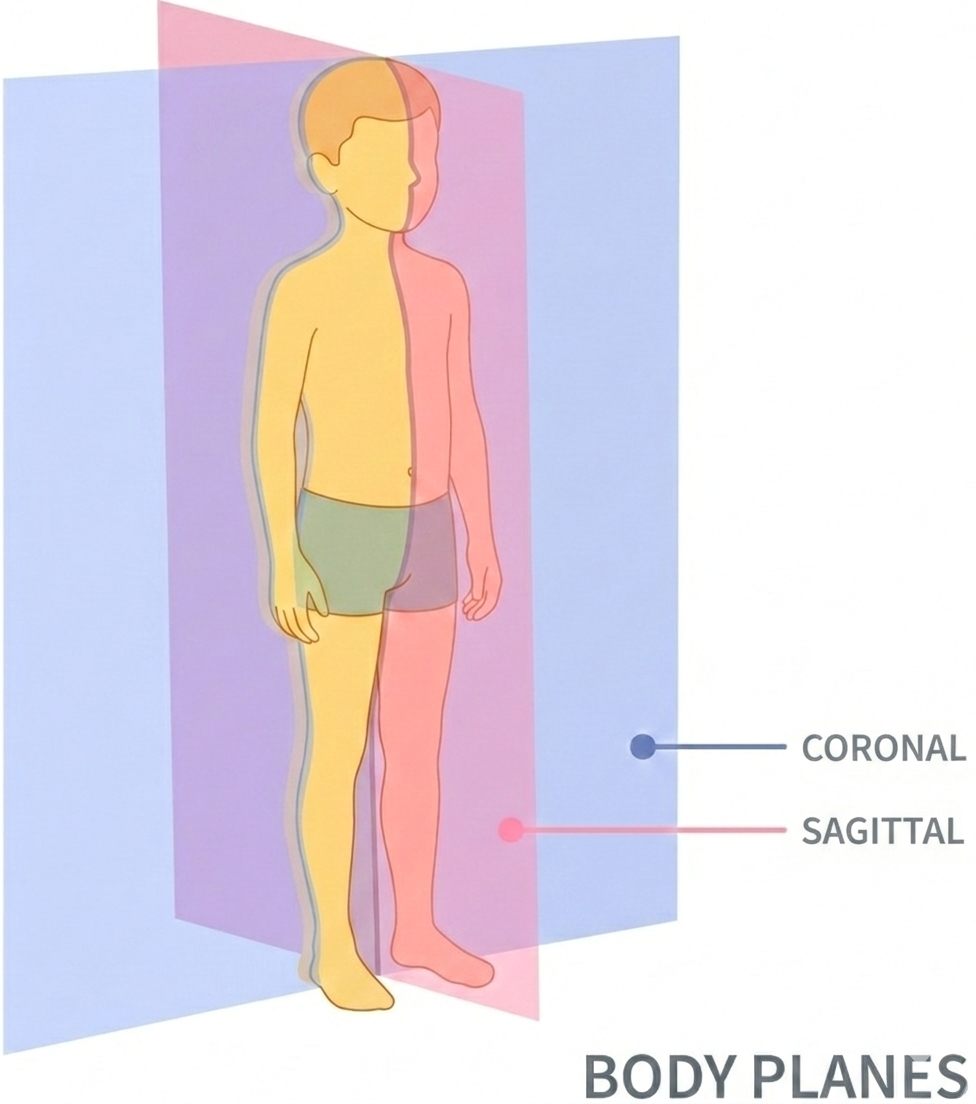}
        \end{minipage}
        
    \end{minipage}
    \hfill
    \begin{minipage}[c]{0.74\textwidth}
        \centering
        
        \begin{minipage}[c]{\textwidth}
            \centering
            \includegraphics[width=0.99\textwidth]{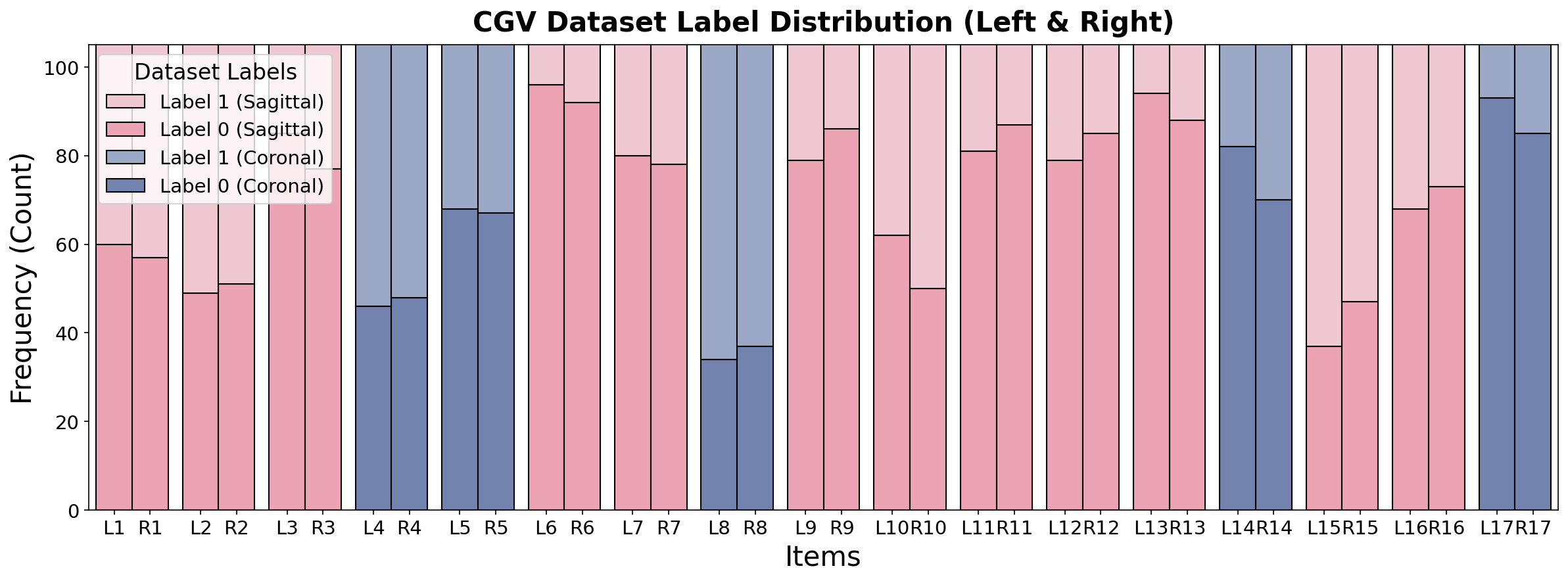}
        \end{minipage}
    \end{minipage}
    \caption{\textbf{The Overview of our CGV Dataset.} The CGV dataset comprises 110 pediatric patients, with each evaluation covering a total of $2\times17$ EVGS scoring items (17 items per limb) across two body planes. \textit{Left:} The child's body planes. \textit{right:} The label distribution of the CGV dataset.}
    \label{fig:CGV_overview}
\end{figure}

We introduce a dataset named CGV for the development of children's gait modeling. This is the first open-sourced children's gait video dataset. Before the data annotation and model design, the IRB approval is obtained from affiliated hospitals. The dataset contains 339,236 frames (1,185 videos) of size $1920\times 1080$ (2K) with 17 EVGS~\cite{read2003edinburgh} sub-item annotations per limb and the diagnosis results of multiple gait abnormalities, such as Cerebral Palsy (CP), Traumatic Brain Injury (TBI), Developmental Dysplasia of the Hip (DDH), Toe in, Idiopathic Toe Walking (ITW). The videos were recorded at a children's hospital in Asia using smartphone cameras and action cameras, positioned simultaneously to capture sagittal and coronal views. \Cref{fig:visualization} shows examples of videos and annotations. In total, 110 patients participated in the data collection. In some cases, multiple recordings per view were available. The dataset provides detailed annotations per frame and per video, including the subject’s body bounding box (detected with SAM 3~\cite{carion2025sam} and manually selected by human annotators), the 2D human keypoints (detected with Sapiens-2B~\cite{khirodkar2024sapiens} and manually adjusted by human annotators), and the EVGS sub-items (annotated by an experienced pediatricians in the author team and reviewed by a senior pediatrician author with 40 years of clinical experience), the diagnosis results tracing from the patient's follow-up record. \Cref{fig:CGV_overview} shows the EVGS scores distribution among all patients. \Cref{sec:details} shows the demographic statistics of the CGV dataset.

To capture the dataset, a primary camera is positioned at the terminus of an 8-meter walkway to record the coronal (frontal and posterior) view. A secondary camera is oriented orthogonally, facing the center of the walkway, to capture the sagittal (lateral) view.  This lateral camera is positioned at a sufficient distance to ensure its field of view encompasses the middle four meters of the trial space. This specific distance is calibrated to guarantee the capture of 2-3 complete gait cycles (strides) per subject. \Cref{tab:dataset_item} details the 17 fine-grained gait parameters annotated in the CGV dataset, which strictly adhere to the EVGS reference guide to ensure high clinical validity. While the standard EVGS protocol employs a three-point ordinal scale (0: normal, 1: moderate deviation, 2: severe deviation) for each parameter, the natural distribution of pediatric gait pathologies inherently results in a severe class imbalance, particularly for the most extreme deviations (score 2). To mitigate this imbalance and establish a robust computational benchmark, we binarize the assessment by merging scores 1 and 2. Consequently, the prediction for each fine-grained gait parameter is formulated as a binary classification task (see \Cref{fig:CGV_overview}) discriminating between ``typical'' and ``atypical'' gait patterns. Detailed EVGS scoring criteria are provided in~\Cref{sec:evgs}. \looseness=-1

\noindent \textbf{Expert Agreement.} To further verify the agreements between experts and the objectiveness of our clinical annotations, we add an additional annotation round from another pediatrician and calculate the intra-class correlation coefficient (ICC) with ours, achieving $\mathrm{ICC}=0.93$. We also invite the pediatrician as a human baseline, achieving an average scoring accuracy of 93.8\%. 

\noindent \textbf{Ethical Considerations and Data Privacy.} Consistent with established ethical frameworks for pediatric facial, pose, and behavioral datasets~\cite{nojavanasghari2016emoreact,tafasca2023childplay,medvedev2024young,rehg2013decoding,cao2022aggpose, huang2023posture}, we have implemented a multi-layered protocol to ensure participant protection. The CGV dataset was approved as a retrospective study by the Institutional Review Board (IRB) of our affiliated clinical institution (approval number: 202106202). Our commitment to participant privacy and data security is reflected in the following measures: \textbf{(1) Irreversible De-identification:} To protect the identities of the children, all original video data underwent an irreversible obfuscation process. We employed an automated preprocessing pipeline utilizing RetinaFace~\cite{deng2020retinaface} for face detection and SAM 3~\cite{carion2025sam} for precise segmentation, applying a mosaic filter to all identifiable regions. To ensure 100\% efficacy, the output was manually verified by two independent human annotators. \textbf{(2) Restrictive Licensing and AI Governance:} The dataset is released under the CC BY-NC 4.0 license. The public avliable data is the children's pose sequence. Furthermore, we mandate that video data and annotation from this repository may not be used as training material for large-scale foundation models without explicit, separate authorization, safeguarding against unauthorized generative use.

\noindent \textbf{Experimental Setup.} 
To prevent identity leakage and shortcut bias, we implement a strict object-level data split based on the unique \textbf{Patient ID} in the experiment. The test set is randomly selected, and we ensure all positive samples and negative sample nearly balanced. All videos and derived gait cycles for any given patient belong exclusively to either the training or the test set. Finally, the ratio of patients in the training set to those in the test set is 6:1. Then, we evaluate all models on the CGV test split on an item-wise basis. We report the per-item percentage accuracy and per-limb average accuracy across all 34 bilateral scoring items to provide a holistic measure of clinical reliability. All experiments are conducted on a cluster equipped with 10 NVIDIA L40S GPUs. For model training in \Cref{sec:vlm_gait} and \Cref{sec:video_gait}, we use AdamW as the optimizer with $\eta=10^{-4}$ and a weight decay of 0.05. We employ a linear warm-up schedule over the first 5\% of the total training epochs to stabilize early optimization. The batch size is configured to 8 per GPU. 

%% file: tabs/items.tex
\begin{table}[!t]
    \centering
    \caption{\textbf{17 Scoring Items in CGV Referring to EVGS.}}
    \label{tab:dataset_item}
    \resizebox{0.8\linewidth}{!}{ 
    \begin{tabular}{lllccc}
        \toprule
        \textbf{No.} & \textbf{Gait Pattern} & \textbf{Abbr.} & \textbf{Score Scale} & \textbf{View} & \textbf{\#Num of Video} \\
        \midrule
        1 & Initial Contact in Stance & IC & 0, 1, 2 & Sagittal & 647 \\
        2 & Heel Lift in Stance & HL & 0, 1, 2 & Sagittal & 647 \\
        3 & Max Ankle Dorsiflexion in Stance & SAD & 0, 1, 2 & Sagittal & 647 \\
        4 & Hind-foot Varus/Valgus in Stance & HVV & 0, 1, 2 & Coronal & 261 \\
        5 & Foot Rotation in Stance & FRT & 0, 1, 2 & Coronal & 277 \\
        6 & Foot Clearance in Swing & FCL & 0, 1 & Sagittal & 647 \\
        7 & Max Ankle Dorsiflexion in Swing & WAD & 0, 1, 2 & Sagittal & 647 \\
        8 & Knee Progression Angle in Mid-Stance & KPA & 0, 1, 2 & Coronal & 277 \\
        9 & Peak Knee Extension in Stance & KEX & 0, 1, 2 & Sagittal & 647 \\
        10 & Knee Position in Terminal Swing & KPS & 0, 1, 2 & Sagittal & 647 \\
        11 & Peak Knee Flexion in Swing & KFX & 0, 1, 2 & Sagittal & 647 \\
        12 & Peak Hip Extension in Stance & HEX & 0, 1, 2 & Sagittal & 647 \\
        13 & Peak Hip Flexion during Swing & HFX & 0, 1, 2 & Sagittal & 647 \\
        14 & Pelvic Obliquity at Mid-Stance & POB & 0, 1, 2 & Coronal & 277 \\
        15 & Pelvic Rotation at Mid-Stance & PRT & 0, 1, 2 & Coronal & 277 \\
        16 & Peak Sagittal Trunk Position in Stance & TSG & 0, 1, 2 & Sagittal & 647 \\
        17 & Maximum Trunk Lateral Shift & TLT & 0, 1, 2 & Coronal & 277 \\
        \bottomrule
    \end{tabular}
    }
    % \vspace{-3mm}
\end{table}

%% file: sec/5_decoding.tex
\section{Benchmarking Children's Gait Analysis}
\label{sec:benchmarking}

\noindent \textbf{Baselines.} We evaluate a wide range of zero-shot multimodal LLMs like Gemini 3 Pro~\cite{google2025gemini3pro}, GPT-5.2~\cite{openai2025gpt5.2}, Qwen3-VL-235B~\cite{li2026qwen3}, GLM4.6V~\cite{zeng2025glm}, Qwen3.5-9B~\cite{qwen35blog}, and InternVL3-8B~\cite{zhu2025internvl3}, validating if these SoTA methods can resolve the children's gait analysis problem. All models take as input uniformly sampled $T=16$ frames from the center window of the downsampled 30 FPS video and the prompt \textbf{P}. \Cref{sec:prompts} shows the additional details of $\mathbf{P}$. \looseness=-1

\noindent \textbf{Experimental Results.} As demonstrated in~\Cref{tab:zero_shot_results}, all listed MLLMs fail to effectively resolve the required clinical nuances, with the average accuracy of each limb range 50\% to 60\%, which is only marginally above random guessing for binary scoring tasks. This performance gap indicates that, although these models exhibit strong general visual–language reasoning abilities, they struggle to capture subtle kinematic deviations and fine-grained temporal dynamics that are critical for clinical gait assessment. More specifically, zero-shot MLLMs tend to correctly identify visually obvious abnormalities like Peak Hip Flexion during Swing (HFX) with an average accuracy of 61\%, but frequently misclassify subtle impairments such as Max Ankle Dorsiflexion in Stance (SAD) with an average accuracy of 54\%. The results suggest that pre-trained MLLMs are biased toward semantic-level understanding rather than quantitative biomechanical interpretation. Additionally, the lack of task-specific alignment between medical scoring criteria and general-purpose language supervision further limits their discriminative capacity. These findings highlight the necessity of domain-adaptive fine-tuning and structured supervision to bridge the gap between generic multimodal reasoning and clinically grounded gait analysis.

\input{tabs/zero_shot_results}

\section{Decoding Children's Gait via Vision-Language Models}
\label{sec:vlm_gait}

We first validate whether Video Vision-Language Models (VLMs) are the best solution for children's gait visual analysis. We propose a multimodal reasoning framework based on a fine-tuned Qwen3-VL-4B~\cite{li2026qwen3} to map pediatric gait patterns to all fine-grained items in CGV, denoted as Qwen3-VL-ChildGait.

\noindent \textbf{Multimodal Representation.} To capture the complex kinematics as we described, each video frame at step $t$ is represented as a composite input $\mathcal{F}_t = \{ \mathbf{I}_t, \mathbf{K}_t, \mathbf{M}_t \}$, where $\mathbf{I}_t$ denotes the original RGB frame, $\mathbf{K}_t$ denotes the skeletal keypoints, and $\mathbf{M}_t$ denotes the instance segmentation mask. This multi-view visual evidence allows the model's transformer layers to attend to both fine-grained joint angles and global body morphology, filtering out clinical background noise.

\noindent \textbf{Clinical Goal-Oriented Fine-tuning.} We formulate gait analysis as a sequence-to-label task. Given a video sequence $\mathbf{F} = \{\mathcal{F}_t\}_{t=1}^T$ which is downsampled to 30 FPS and where we randomly crop $M\!=\!4$ temporal windows and uniformly sample $T\!=\!16$ per window during training and uniformly sample $T\!=\!16$ in the single central temporal window during testing, the model is fine-tuned to predict a categorical gait score $y \in \mathcal{Y}$. The training objective minimizes the negative log-likelihood of the target tokens:

\begin{equation}\mathcal{L}_{\text{vlm}}(\boldsymbol{\theta}) = - \sum{j} \log p_{\boldsymbol{\theta}}(w_j \mid w_{<j}, \mathbf{F}, \mathbf{P}),
\end{equation}

\input{tabs/vlm_results}

where $\mathbf{P}$ is a task-specific prompt instructing the model to synthesize visual trajectories into gait quality ratings. This approach uses the VLM’s pre-trained spatial reasoning to identify sub-second deviations that traditional adult-centric models often miss. \Cref{sec:prompts} shows the additional details of $\mathbf{P}$. \looseness=-1

\noindent \textbf{Experimental Results.} We compare zero-shot Qwen3-VL-4B~\cite{li2026qwen3} with the fine-tuned Qwen3-ChildGait on the CGV test split. As shown in~\Cref{tab:vlm_results}, fine-tuned Qwen3-VL-4B has no significant improvement in performance compared to the base model, with the average accuracy of left and right limbs increasing 2\% and 1\%, respectively.
While certain scoring items exhibit little gains in accuracy, such as Initial Contact in Stance (IC) with +11.5\%, some of the other items show decreased performance after fine-tuning, such as Hind-foot Varus/Valgus in Stance (HVV) with -1.5\%, resulting in a mixed overall average. The counterintuitive result demonstrates that VLMs do not perform well on the subtle task of child gait recognition, which aligns with recent findings that VLMs face fundamental limitations in fine-grained action recognition and dynamic spatiotemporal interactions~\cite{li2025potential, zhou2025vlm4d}. The core of VLMs' training lies in modeling high-dimensional discrete tokens. By performing causal language modeling across trillions of text corpora, it fundamentally captures the statistical distribution of human knowledge within logical and semantic spaces. Therefore, knowledge compressed from human language results cannot be generalized to motion recognition~\cite{upadhyay2025time}, much less to the more refined task of gait recognition. 

\section{Decoding Children's Gait via Video-based Models}
\label{sec:video_gait}

\begin{figure}[!t]
    \centering
    \includegraphics[width=0.99\linewidth]{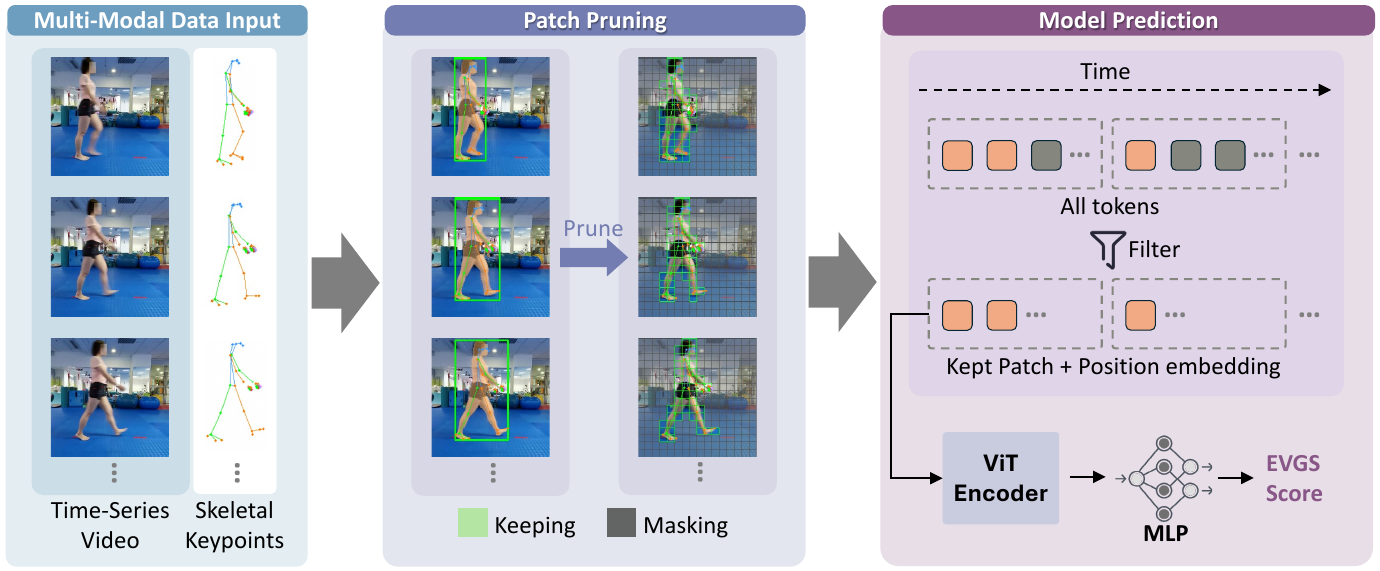}
    \caption{\textbf{ChildGait-Video Framework.} We render the skeletal keypoints on the original RGB frame as the token-level kinematic prompts and then perform mask-guided patch pruning to force the model to focus on the foreground patient. The processed input is then passed into the model to predict EVGS scoring items.}
    \label{fig:framework}
\end{figure}

The failure of Video VLMs prompts us to rethink whether existing video-based foundation and gait analysis models can resolve this problem after fully fine-tuning in CGV.

\subsection{Fine-tuning Video Foundation and Gait Analysis Models}
\label{sec:finetune}

We establish our video model baseline by directly fine-tuning VideoMAE v2~\cite{wang2023videomae}, a state-of-the-art video foundation model. It is pre-trained via masked autoencoding on the UnlabeledHybrid dataset to capture generic spatiotemporal knowledge, and subsequently fine-tuned with supervision on Kinetics-710~\cite{li2022uniformerv2} to acquire human motion recognition capabilities.

In addition to the video foundation model baseline, we also evaluate a wide range of representative gait analysis baselines. These include the current state-of-the-art (SoTA) model BiggerGait~\cite{ye2025biggergait}, as well as its precursors: GaitSet~\cite{chao2019gaitset, chao2021gaitset}, GaitPart~\cite{fan2020gaitpart}, GaitGL~\cite{lin2021gait}, GaitBase~\cite{fan2023opengait}, SwinGait~\cite{fan2023exploring}, and DeepGaitV2~\cite{fan2023exploring}. These models are first pre-trained on various large-scale datasets (\eg, CCPG~\cite{li2023depth}, Gait3D~\cite{zheng2022gait}, Gait3D-Parsing~\cite{zheng2023parsing}, SUSTech1K~\cite{shen2023lidargait}, GREW~\cite{zhu2021gait}, OUMVLPP~\cite{takemura2018multi}) and subsequently fine-tuned on the CGV dataset.

\noindent \textbf{Video Downsample and Clip.} To standardize the temporal input distribution and align it with the model's expectation, the raw video stream is first downsampled to a fixed frame rate of 30 FPS, ensuring that the physical time stride between adjacent frames remains constant across the dataset. For temporal augmentation during training, we randomly crop $M\!=\!4$ temporal windows and uniformly sample $T\!=\!16$ frames per window. During testing, a single $T\!=\!16$ frame sequence is uniformly extracted from the video's central temporal window for deterministic evaluation. Thus, the visual input is formally defined as $\mathbf{V} \in \mathbb{R}^{T \times 3 \times H \times W}$, where $H$ and $W$ denote the spatial resolution, which is then flattened to patches. This ensures a sufficient temporal receptive field to capture a complete gait cycle phase without overwhelming the computational budget.

\noindent \textbf{Task-Specific Architecture Adaptation.} Let the output from the final transformer block be denoted as $\textbf{X}\in \mathbb{R}^{B \times N \times D}$, where $B$ is the batch size, $N$ is the total number of spatiotemporal patch tokens, and $D$ is the embedding dimension. \looseness=-1

To aggregate the global context, we perform spatiotemporal global average pooling over all $N$ patch tokens, yielding a compact clip-level feature of shape $B \times D$. Finally, a newly initialized linear classification head projects the normalized features into a two-dimensional logit vector:

\begin{equation}
    \mathbf{L} = \mathbf{W}_{\text{head}} \, \left( \frac{1}{N} \sum_{i=1}^{N} \mathbf{x}_i \right) + \mathbf{b}_{\text{head}},
\end{equation}
where $\mathbf{x}_i$ represents the $i$-th token, and $\mathbf{W}_{head} \in \mathbb{R}^{2 \times D}$. The final prediction is obtained via the $\arg\max$ operation over the two logits, and the model is optimized end-to-end using the Soft Target Cross-Entropy loss:

\begin{equation}
    \mathcal{L}_{\text{video}}(\boldsymbol{\theta}) = -\sum_{k\in\{0, 1\}}q_k\log \left( \text{Softmax}(\mathbf{L}(\theta))_k\right),
\end{equation}
where $\textbf{q} = (\lambda, 1-\lambda), \lambda \in [0, 1]$ is the proportion of positive/negative samples.\looseness=-1

\subsection{Designing ChildGait-Video Baseline Aligning with Clinical Intuition}

\input{tabs/video_results}

Traditional gait analysis methods mostly follow a multi-stage pipeline: pose estimation for keypoint extraction, geometric calculation of joint angles, and rule-based scoring. However, this geometry-centric paradigm suffers from two major limitations. First, compressing high-dimensional video data into sparse skeletal representations inevitably discards rich appearance and texture information. Second, multi-stage pipelines typically evaluate specific joint angles in isolation, neglecting the inherent kinematic synergy of human movement.

As shown in~\Cref{fig:framework}, to address these limitations, we propose \textbf{ChildGait-Video}, an end-to-end model that directly maps visual features to EVGS items. ChildGait-Video uses the same architecture of vision encoder as VideoMAE v2~\cite{wang2023videomae}, followed by an MLP head to predict EVGS items. This end-to-end design inherently mitigates multi-stage error accumulation and uses the strong spatiotemporal representation capabilities of video foundation models to capture the global motion patterns and inter-dependencies among EVGS items. 

To unleash the potential of the model for pediatric gait analysis, we propose a two-stage adaption paradigm. We first pre-train ChildGait-Video on Kinetics-710~\cite{li2022uniformerv2} to acquire human motion recognition capability, and then supervised fine-tuning the model on our proposed CGV dataset with \textbf{Token-Level Kinematic Prompting} and \textbf{Mask-Guided Patch Pruning} to further make the model learn the subtle gait anomalies as well as focus on the gait transition. 

\noindent \textbf{Anatomical Alignment via Token-Level Kinematic Prompting.}
Although pre-trained on Kinetics-710, the model still lacks an explicit understanding of the underlying anatomical priors of children, which is a critical requirement for identifying subtle gait anomalies. To bridge this semantic gap, we introduce skeletal keypoints as token-level kinematic prompts.

We directly render the extracted joint coordinates $\mathbf{K}_t$ and their connectivity graph onto the original RGB frame $\mathbf{I}_t$. Let $\mathcal{R}(\cdot)$ denote this rendering operation. The prompted input frame is formulated as $\hat{\mathbf{I}}_t = \mathcal{R}(\mathbf{I}_t, \mathbf{K}_t)$. During the flattening process in Vision Transformer (ViT) encoder~\cite{dosovitskiy2020image}, the image is divided into patches that encompass the rendered skeletal keypoints and connections, inherently encapsulating both local anatomical topology and the original appearance features. When projected into the embedding space, these specific patches act as token-level kinematic prompts mixed with standard visual tokens. 

\noindent \textbf{Background Suppression via Mask-Guided Patch Pruning.}
Gait videos are often cluttered with irrelevant background noises, which easily introduce biases during fine-tuning. Inspired by the inherent sparsity of the MAE architecture, we utilize instance segmentation to extract binary subject masks $\mathbf{M} \in \{0, 1\}^{H \times W}$. Rather than employing random masking, we perform deterministic spatial pruning guided by these masks. 

Specifically, a processed RGB frame $\hat{\mathbf{I}}_t$ is divided into a grid of non-overlapping patches. A patch is retained only if its corresponding region in $\mathbf{M}$ contains a sufficient proportion of foreground pixels. Let $\mathbf{E}_\text{v} \in \mathbb{R}^{N_{\text{foreground}} \times D}$ denote the embedded representations of the retained foreground patches, where~$N_{\text{foreground}} \ll N_{\text{total}}$. This mask-guided token dropping strategy not only explicitly eliminates background noise, forcing the transformer's self-attention to strictly allocate its representational capacity to the subject's gait, but also substantially reduces the computational overhead during fine-tuning.

\noindent \textbf{Experimental Results.} As shown in~\Cref{tab:video_results}, gait analysis models also struggle to perform precise classification in a clinical setting, yielding low average accuracies that range from just 45\% to 56\%. Notably, SwinGait pre-trained on Gait3D and GaitBase pre-trained on GREW achieve marginally better results, peaking at average accuracies of 54\% and 56\% for the left and right limbs, respectively. Although being the current SoTA gait analysis model, BiggerGait still fails to do the classification precisely with low average accuracies of 54\% and 53\%, respectively, for left and right limbs. We attribute their low accuracy to the inherent design objectives of traditional gait recognition networks, which focus on extracting global spatiotemporal representations to distinguish identities, thereby neglecting the fine-grained, localized kinematic details essential for clinical scoring. Furthermore, there exists a domain gap between the healthy adult subjects in the pre-training datasets and the distinct pathological patterns of children's gait in the CGV dataset, which limits their feature representation and generalization capabilities. 

On the other hand, fine-tuned VideoMAE v2 has surpassed all previous baselines in almost every scoring item with average accuracies of 69\% and 72\%, increased by 15\% and 19\% compared to BiggerGait.

Finally, ChildGait-Video achieves SoTA performance across all baselines, reaching a highest accuracy range of $70\% \sim 93\%$ in all 34 scoring items. It gains average accuracy improvements of 15\% and 12\% for left and right limbs compared to the fine-tuned VideoMAE v2. The average F1-scores also achieve a high value of 0.83. This SoTA performance can be attributed to the interaction within the self-attention mechanism. We also evaluate skeleton baselines in~\Cref{sec:skeleton}. \looseness=-1

\noindent \textbf{Statistical Analysis.} We employ McNemar's test on ChildGait-Video and fine-tuned VideoMAE v2. Let $b$ denote the number of data that ours gets correct while the baseline gets wrong, and $c$ the opposite. We compute the $p$-value using a binomial test under the null hypothesis that both models have equal accuracy, and get $(b,c)=(28, 6), p=2.3\times10^{-4} < 0.001$, demonstrating that the performance improvement is highly statistically significant.

\begin{figure}[!t]
    \centering
    \includegraphics[width=0.99\linewidth]{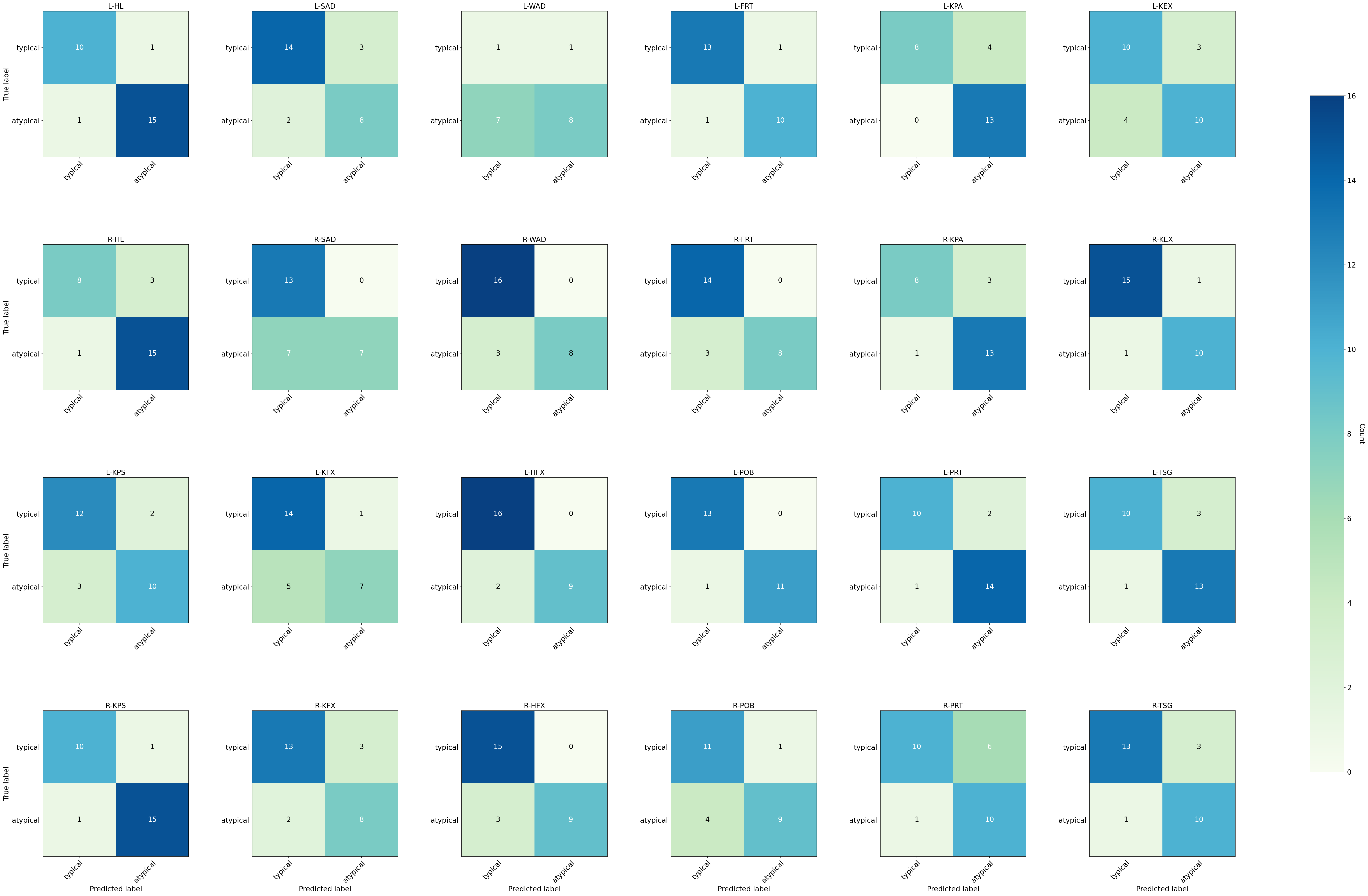}
    \caption{\textbf{Visualization of Confusion Matrices.} We visualize confusion matrices covering both sagittal and coronal views of scoring items in EVGS.}
    \label{fig:confusion}
\end{figure}

\noindent \textbf{Confusion Matrices.} We also visualize confusion matrices covering both sagittal and coronal views of scoring items in EVGS. As depicted in~\Cref{fig:confusion}, our framework maintains high sensitivity for both sagittal view items and coronal view items. The low off-diagonal error rate for these items suggests that our method effectively regularizes the self-attention mechanism to focus on gait kinematics.

\subsection{Ablation Study on Frames}

The temporal resolution of the input sequence determines the model's capacity to capture the dynamic evolution of gait pathologies. We investigate the impact of the number of sampled frames $T \in \{8, 16, 32\}$ while maintaining a constant sampling rate of 30 FPS through training and evaluating on item $\textbf{L/R-IC}$. All variants leverage the proposed mask-guided pruning and visual prompting with ChildGait-Video. We also conduct a module ablation study in~\Cref{sec:module}.

As shown in~\Cref{tab:frame_ablation}, increasing the frame count from 8 to 16 yields a significant performance gain with an increase of 12.9\%. The performance is further increased to 90.7\% if the number of frames is increased to 32. We attribute this to the fact that 8 frames are insufficient to encompass a meaningful kinematic transition in pediatric gait. In contrast, 16 frames effectively capture critical sub-phases of the gait cycle. Interestingly, further increasing the temporal window to 32 frames results in only a slight improvement in performance of 3.7\%, as computational complexity grows quadratically, although the field of view has doubled. \looseness=-1

\input{tabs/ablation.tex}

%% file: tabs/zero_shot_results.tex
\begin{table*}[!t]
    \centering
    \caption{\textbf{Zero-shot Quantitative Evaluation of Mainstream MLLMs.} All reported values are \textbf{percentages (\%)}. We report 17 scoring items for the \textbf{Left (L-)} limb (top) and \textbf{Right (R-)} limb (bottom). The detailed definition of each item is shown in~\Cref{tab:dataset_item}. L/R-AVG indicates the average accuracy over all scoring items of the left/right limb.}
    \label{tab:zero_shot_results}
    \setlength{\tabcolsep}{3.3pt}
    
    % --- Table: Left Limb ---
    \resizebox{0.8\linewidth}{!}{%
        \begin{tabular}{@{} l *{18}{c} @{}}
            \toprule
            \textbf{Method} & \mrot{L-IC} & \mrot{L-HL} & \mrot{L-SAD} & \mrot{L-WAD} & \mrot{L-HVV} & \mrot{L-FRT} & \mrot{L-FCL} & \mrot{L-KPA} & \mrot{L-KEX} & \mrot{L-KPS} & \mrot{L-KFX} & \mrot{L-HEX} & \mrot{L-HFX} & \mrot{L-POB} & \mrot{L-PRT} & \mrot{L-TSG} & \mrot{L-TLT} & \mrot{L-AVG}\\
            \midrule
            \rule{0pt}{0ex}Gemini 3 Pro~\cite{google2025gemini3pro} & 52 & 48 & 41 & 59 & 39 & 57 & 52 & 46 & 67 & 48 & 52 & 63 & 59 & 46 & 41 & 70 & 54 & 53 \\
            GPT-5.2~\cite{openai2025gpt5.2} & 56 & 63 & 44 & 59 & 42 & 54 & 56 & 54 & 48 & 44 & 59 & 63 & 59 & 54 & 44 & 56 & 54 & 53 \\
            Qwen3-VL-235B~\cite{li2026qwen3} & 59 & 44 & 70 & 48 & 58 & 57 & 56 & 46 & 59 & 48 & 56 & 56 & 59 & 50 & 48 & 52 & 57 & 54 \\
            GLM4.6V~\cite{zeng2025glm} & 56 & 48 & 67 & 41 & 39 & 57 & 63 & 46 & 56 & 48 & 52 & 59 & 63 & 50 & 37 & 52 & 57 & 52 \\
            Qwen3.5-9B~\cite{qwen35blog} & 48 & 44 & 63 & 44 & 35 & 57 & 59 & 39 & 52 & 44 & 56 & 56 & 59 & 54 & 41 & 48 & 54 & 50 \\
            InternVL3-8B~\cite{zhu2025internvl3} & 52 & 44 & 63 & 44 & 42 & 61 & 59 & 57 & 52 & 44 & 56 & 56 & 59 & 54 & 41 & 48 & 68 & 53 \\

            \bottomrule
        \end{tabular}
    }
    
    \vspace{2mm}
        
    % --- Table: Right Limb + Avg Acc ---
    \resizebox{0.8\linewidth}{!}{%
        \begin{tabular}{@{} l *{18}{c} r @{}}
            \toprule
            \textbf{Method} & \mrot{R-IC} & \mrot{R-HL} & \mrot{R-SAD} & \mrot{R-WAD} & \mrot{R-HVV} & \mrot{R-FRT} & \mrot{R-FCL} & \mrot{R-KPA} & \mrot{R-KEX} & \mrot{R-KPS} & \mrot{R-KFX} & \mrot{R-HEX} & \mrot{R-HFX} & \mrot{R-POB} & \mrot{R-PRT} & \mrot{R-TSG} & \mrot{R-TLT} & \mrot{R-AVG} \\
            \midrule
            \rule{0pt}{0ex}Gemini 3 Pro~\cite{google2025gemini3pro} & 59 & 59 & 48 & 44 & 42 & 57 & 48 & 46 & 52 & 52 & 59 & 67 & 70 & 54 & 59 & 63 & 64 & 55  \\
            GPT-5.2~\cite{openai2025gpt5.2} & 63 & 59 & 63 & 52 & 46 & 61 & 59 & 46 & 52 & 63 & 52 & 44 & 67 & 50 & 59 & 44 & 64 & 56 \\
            Qwen3-VL-235B~\cite{li2026qwen3} & 52 & 59 & 48 & 67 & 58 & 61 & 70 & 43 & 67 & 56 & 56 & 70 & 59 & 46 & 67 & 63 & 57 & 59 \\
            GLM4.6V~\cite{zeng2025glm} & 44 & 56 & 48 & 59 & 42 & 64 & 56 & 39 & 67 & 48 & 59 & 63 & 63 & 50 & 52 & 59 & 61 & 55  \\
            Qwen3.5-9B~\cite{qwen35blog} & 52 & 48 & 48 & 59 & 42 & 57 & 56 & 46 & 59 & 41 & 59 & 56 & 56 & 46 & 59 & 59 & 54 & 53  \\
            InternVL3-8B~\cite{zhu2025internvl3} & 44 & 48 & 48 & 59 & 46 & 57 & 56 & 57 & 59 & 41 & 59 & 56 & 56 & 46 & 59 & 59 & 64 & 54 \\

            \bottomrule
        \end{tabular}
    }
\end{table*}

%% file: tabs/vlm_results.tex
\begin{table*}[!t]
    \centering
    \caption{\textbf{Comparison Between Fine-tuned (Or Not) Qwen3-VL-4B.} All reported values are \textbf{percentages (\%)}. We report 17 scoring items for the \textbf{Left (L-)} limb (top) and \textbf{Right (R-)} limb (bottom). The detailed definition of each item is shown in~\Cref{tab:dataset_item}. L/R-AVG indicates the average accuracy over all scoring items of the left/right limb.}
    \label{tab:vlm_results}
    \scriptsize
    \setlength{\tabcolsep}{3.3pt}
    
    % --- Table: Left Limb ---
    \resizebox{0.9\linewidth}{!}{%
    \begin{tabular}{@{} l *{18}{c} @{}}
        \toprule
        \textbf{Method} & \mrot{L-IC} & \mrot{L-HL} & \mrot{L-SAD} & \mrot{L-WAD} & \mrot{L-HVV} & \mrot{L-FRT} & \mrot{L-FCL} & \mrot{L-KPA} & \mrot{L-KEX} & \mrot{L-KPS} & \mrot{L-KFX} & \mrot{L-HEX} & \mrot{L-HFX} & \mrot{L-POB} & \mrot{L-PRT} & \mrot{L-TSG} & \mrot{L-TLT} & \mrot{L-AVG} \\
        \midrule

        \rule{0pt}{0ex}Qwen3-VL-4B~\cite{li2026qwen3} & 52 & 44 & 63 & 44 & 42 & 54 & 59 & 50 & 52 & 44 & 56 & 56 & 59 & 46 & 41 & 48 & 54 & 51 \\
        Qwen3-VL-ChildGait & 63 & 56 & 59 & 48 & 39 & 57 & 52 & 46 & 52 & 56 & 52 & 56 & 63 & 50 & 41 & 48 & 57 & 53 \\

        \bottomrule
    \end{tabular}
    }
 \vspace{2mm}
    
    % --- Table: Right Limb + Avg Acc ---
    \resizebox{0.9\linewidth}{!}{%
    \begin{tabular}{@{} l *{18}{c} r @{}}
        \toprule
        \textbf{Method} & \mrot{R-IC} & \mrot{R-HL} & \mrot{R-SAD} & \mrot{R-WAD} & \mrot{R-HVV} & \mrot{R-FRT} & \mrot{R-FCL} & \mrot{R-KPA} & \mrot{R-KEX} & \mrot{R-KPS} & \mrot{R-KFX} & \mrot{R-HEX} & \mrot{R-HFX} & \mrot{R-POB} & \mrot{R-PRT} & \mrot{R-TSG} & \mrot{R-TLT} & \mrot{R-AVG}\\
        \midrule

        \rule{0pt}{0ex}Qwen3-VL-4B~\cite{li2026qwen3} & 44 & 48 & 48 & 59 & 42 & 46 & 56 & 36 & 59 & 41 & 59 & 56 & 56 & 54 & 59 & 59 & 54 & 52\\
        Qwen3-VL-ChildGait & 56 & 52 & 56 & 67 & 42 & 61 & 56 & 43 & 59 & 59 & 67 & 56 & 56 & 46 & 59 & 59 & 57 & 53 \\

        \bottomrule
    \end{tabular}
    }
\end{table*}

%% file: tabs/video_results.tex
\begin{table*}[!t]
    \centering
    \caption{\textbf{Quantitative Evaluation of Video Analysis Models.} All reported values are \textbf{percentages (\%)}. We report 17 scoring items for the \textbf{Left (L-)} limb (top) and \textbf{Right (R-)} limb (bottom). The detailed definition of each item is shown in~\Cref{tab:dataset_item}. L/R-AVG indicates the average accuracy over all scoring items of the left/right limb. L/R-F1 indicates the average F1-score across all items of the left/right limb.}
    \label{tab:video_results}
    \setlength{\tabcolsep}{3.3pt}
    
    % --- Table: Left Limb ---
    \resizebox{0.99\linewidth}{!}{%
        \begin{tabular}{@{} ll *{19}{c} @{}}
            \toprule
            \textbf{Method} & \textbf{Pretrain} & \mrot{L-IC} & \mrot{L-HL} & \mrot{L-SAD} & \mrot{L-WAD} & \mrot{L-HVV} & \mrot{L-FRT} & \mrot{L-FCL} & \mrot{L-KPA} & \mrot{L-KEX} & \mrot{L-KPS} & \mrot{L-KFX} & \mrot{L-HEX} & \mrot{L-HFX} & \mrot{L-POB} & \mrot{L-PRT} & \mrot{L-TSG} & \mrot{L-TLT} & \mrot{L-AVG} & \mrot{L-F1} \\
            \midrule
            \rule{0pt}{0ex}GaitSet~\cite{chao2019gaitset, chao2021gaitset} & Gait3D-Parsing~\cite{zheng2023parsing} & 48 & 44 & 48 & 56 & 54 & 43 & 56 & 54 & 52 & 63 & 48 & 59 & 44 & 50 & 41 & 52 & 57 & 51 & - \\
            GaitPart~\cite{fan2020gaitpart} & Gait3D-Parsing~\cite{zheng2023parsing} & 37 & 44 & 37 & 26 & 30 & 61 & 59 & 50 & 52 & 33 & 63 & 70 & 52 & 50 & 56 & 48 & 43 & 48 & - \\
            GaitGL~\cite{lin2021gait} & Gait3D-Parsing~\cite{zheng2023parsing} & 52 & 56 & 37 & 44 & 39 & 54 & 56 & 54 & 48 & 44 & 37 & 56 & 41 & 50 & 59 & 74 & 32 & 49 & - \\
            GaitBase~\cite{fan2023opengait} & Gait3D~\cite{zheng2022gait} & 56 & 63 & 48 & 44 & 39 & 68 & 22 & 46 & 44 & 48 & 52 & 37 & 41 & 54 & 70 & 44 & 64 & 50 & - \\
            GaitBase~\cite{fan2023opengait} & GREW~\cite{zhu2021gait} & 41 & 70 & 37 & 44 & 69 & 61 & 52 & 57 & 52 & 59 & 44 & 67 & 59 & 50 & 59 & 41 & 43 & 53 & - \\
            GaitBase~\cite{fan2023opengait} & OUMVLP~\cite{takemura2018multi} & 56 & 70 & 37 & 37 & 54 & 50 & 44 & 54 & 52 & 59 & 44 & 41 &41 & 46 & 26 & 52 & 57 & 48 & - \\
            SwinGait~\cite{fan2023exploring} & CCPG~\cite{li2023depth} & 52 & 56 & 30 & 56 & 62 & 57 & 67 & 50 & 48 & 44 & 41 & 44 & 41 & 57 & 59 & 40 & 39 & 50 & - \\
            SwinGait~\cite{fan2023exploring} & Gait3D~\cite{zheng2022gait} & 63 & 44 & 52 & 52 & 39 & 43 & 59 & 50 & 52 & 41 & 44 & 85 & 60 & 57 & 59 & 59 & 57 & 54 & - \\
            SwinGait~\cite{fan2023exploring} & SUSTech1K~\cite{shen2023lidargait} & 63 & 44 & 48 & 44 & 19 & 43 & 59 & 50 & 59 & 56 & 59 & 59 & 52 & 46 & 67 & 48 & 57 & 52 & - \\
            DeepGaitV2~\cite{fan2023exploring} & CCPG~\cite{li2023depth} & 30 & 52 & 63 & 56 & 62 & 54 & 59 & 43 & 74 & 37 & 44 & 44 & 41 & 61 & 30 & 44 & 54 & 50 & - \\
            DeepGaitV2~\cite{fan2023exploring} & Gait3D~\cite{zheng2022gait} & 48 & 67 & 37 & 44 & 62 & 57 & 70 & 32 & 22 & 56 & 44 & 48 & 41 & 50 & 59 & 48 & 61 & 50 & - \\
            DeepGaitV2~\cite{fan2023exploring} & GREW~\cite{zhu2021gait} & 52 & 41 & 41 & 56 & 62 & 57 & 59 & 54 & 52 & 41 & 44 & 56 & 41 & 50 & 33 & 48 & 57 & 50 & - \\
            DeepGaitV2~\cite{fan2023exploring} & OUMVLP~\cite{takemura2018multi} & 48 & 41 & 48 & 52 & 62 & 64 & 48 & 54 & 48 & 44 & 37 & 44 & 59 & 54 & 37 & 52 & 43 & 49 & - \\
            DeepGaitV2~\cite{fan2023exploring} & SUSTech1K~\cite{shen2023lidargait} & 52 & 44 & 37 & 48 & 35 & 43 & 59 & 50 & 48 & 56 & 56 & 56 & 59 & 50 & 59 & 59 & 57 & 51 & - \\
            BiggerGait~\cite{ye2025biggergait} & CCPG~\cite{li2023depth} & 48 & 56 & 63 & 44 & 62 & 43 & 59 & 54 & 52 & 56 & 56 & 56 & 59 & 50 & 59 & 52 & 57 & 54 & 0.53 \\
            \midrule
            VideoMAE v2~\cite{wang2023videomae} & Kinetics-710~\cite{li2022uniformerv2} & 72 & 70 & 70 & 63 & 67 & 68 & 81 & 80 & 60 & 60 & 63 & 67 & 74 & 68 & 80 & 67 & 68 & 69 & 0.70 \\
            \midrule
            \textbf{ChildGait-Video (Ours)} & Kinetics-710~\cite{li2022uniformerv2} & 85 & 92 & 81 & 70 & 88 & 92 & 89 & 84 & 74 & 82 & 78 & 85 & 93 & 89 & 89 & 85 & 76 & 84 & 0.83 \\
            \bottomrule
        \end{tabular}
    }
    
    \vspace{2mm}
        
    % --- Table: Right Limb + Avg Acc ---
    \resizebox{0.99\linewidth}{!}{%
        \begin{tabular}{@{} ll *{19}{c} r @{}}
            \toprule
            \textbf{Method} & \textbf{Pretrain} & \mrot{R-IC} & \mrot{R-HL} & \mrot{R-SAD} & \mrot{R-WAD} & \mrot{R-HVV} & \mrot{R-FRT} & \mrot{R-FCL} & \mrot{R-KPA} & \mrot{R-KEX} & \mrot{R-KPS} & \mrot{R-KFX} & \mrot{R-HEX} & \mrot{R-HFX} & \mrot{R-POB} & \mrot{R-PRT} & \mrot{R-TSG} & \mrot{R-TLT} & \mrot{R-AVG} & \mrot{R-F1} \\
            \midrule
            \rule{0pt}{0ex}GaitSet~\cite{chao2019gaitset, chao2021gaitset} & Gait3D-Parsing~\cite{zheng2023parsing} & 44 & 48 & 63 & 41 & 58 & 39 & 56 & 57 & 59 & 67 & 41 & 59 & 52 & 46 & 30 & 37 & 57 & 50 & - \\
            GaitPart~\cite{fan2020gaitpart} & Gait3D-Parsing~\cite{zheng2023parsing} & 52 & 56 & 52 & 44 & 42 & 43 & 56 & 54 & 41 & 37 & 56 & 67 & 48 & 46 & 41 & 59 & 43 & 49 & - \\
            GaitGL~\cite{lin2021gait} & Gait3D-Parsing~\cite{zheng2023parsing} & 44 & 48 & 52 & 59 & 42 & 57 & 52 & 57 & 41 & 41 & 48 & 59 & 44 & 46 & 37 & 52 & 32 & 48 & - \\
            GaitBase~\cite{fan2023opengait} & Gait3D~\cite{zheng2022gait} & 44 & 41 & 37 & 37 & 42 & 50 & 26 & 43 & 59 & 48 & 59 & 41 & 37 & 50 & 63 & 33 & 50 & 45 & - \\
            GaitBase~\cite{fan2023opengait} & GREW~\cite{zhu2021gait} & 56 & 74 & 52 & 59 & 73 & 57 & 52 & 61 & 59 & 59 & 41 & 63 & 56 & 54 & 41 & 59 & 36 & 56 & - \\
            GaitBase~\cite{fan2023opengait} & OUMVLP~\cite{takemura2018multi} & 67 & 70 & 63 & 56 & 42 & 54 & 52 & 57 & 56 & 44 & 41 & 37 & 59 & 50 & 52 & 41 & 57 & 53 & - \\
            SwinGait~\cite{fan2023exploring} & CCPG~\cite{li2023depth} & 44 & 52 & 56 & 41 & 58 & 68 & 63 & 46 & 48 & 41 & 37 & 41 & 44 & 61 & 41 & 30 & 39 & 48 & - \\
            SwinGait~\cite{fan2023exploring} & Gait3D~\cite{zheng2022gait} & 67 & 44 & 67 & 44 & 42 & 39 & 67 & 54 & 44 & 52 & 59 & 85 & 56 & 61 & 56 & 52 & 57 & 56 & - \\
            SwinGait~\cite{fan2023exploring} & SUSTech1K~\cite{shen2023lidargait} & 78 & 59 & 67 & 63 & 23 & 40 & 56 & 46 & 59 & 59 & 37 & 59 & 52 & 50 & 52 & 56 & 57 & 54 & - \\
            DeepGaitV2~\cite{fan2023exploring} & CCPG~\cite{li2023depth} & 22 & 70 & 48 & 48 & 65 & 50 & 67 & 46 & 56 & 30 & 44 & 44 & 44 & 57 & 37 & 52 & 54 & 49 & - \\
            DeepGaitV2~\cite{fan2023exploring} & Gait3D~\cite{zheng2022gait} & 41 & 74 & 52 & 59 & 58 & 61 & 70 & 29 & 48 & 33 & 41 & 44 & 44 & 46 & 41 & 59 & 46 & 50 & - \\
            DeepGaitV2~\cite{fan2023exploring} & GREW~\cite{zhu2021gait} & 44 & 48 & 52 & 41 & 58 & 61 & 56 & 57 & 48 & 48 & 41 & 56 & 44 & 54 & 48 & 59 & 57 & 51 & - \\
            DeepGaitV2~\cite{fan2023exploring} & OUMVLP~\cite{takemura2018multi} & 56 & 52 & 41 & 41 & 58 & 54 & 52 & 57 & 41 & 41 & 52 & 44 & 56 & 57 & 59 & 41 & 43 & 50 & - \\
            DeepGaitV2~\cite{fan2023exploring} & SUSTech1K~\cite{shen2023lidargait} & 44 & 56 & 52 & 59 & 46 & 39 & 56 & 46 & 59 & 59 & 63 & 56 & 56 & 54 & 41 & 56 & 57 & 53 & - \\
            BiggerGait~\cite{ye2025biggergait} & CCPG~\cite{li2023depth} & 56 & 52 & 48 & 59 & 58 & 39 & 56 & 57 & 59 & 59 & 59 & 56 & 56 & 46 & 41 & 41 & 57 & 53 & 0.51 \\
            \midrule
            VideoMAE v2~\cite{wang2023videomae} & Kinetics-710~\cite{li2022uniformerv2} & 67 & 71 & 61 & 81 & 60 & 68 & 85 & 72 & 78 & 81 & 63 & 67 & 74 & 72 & 70 & 85 & 68 & 72 & 0.71 \\
            \midrule
            \textbf{ChildGait-Video (Ours)} & Kinetics-710~\cite{li2022uniformerv2} & 89 & 85 & 74 & 85 & 83 & 88 & 93 & 80 & 92 & 86 & 81 & 81 & 89 & 80 & 74 & 92 & 76 & 84 & 0.83 \\
            \bottomrule
        \end{tabular}
    }
\end{table*}

%% file: tabs/ablation.tex
\begin{table}[!t]
    \centering
    % \vspace{-5mm}
    \caption{\textbf{Ablation Study on the Number of Input Frames.} We train ChildGait-Video with different numbers of frames and evaluate the performance on the item L/R-IC with the metric Accuracy and F1-Score.}
    \label{tab:frame_ablation}
    \resizebox{0.55\linewidth}{!}{ 
    \begin{tabular}{@{}lccc@{}}
        \toprule
        \textbf{Frames ($T$)} & \textbf{Temporal Span (s)} & \textbf{Accuracy (\%)} & \textbf{F1-Score} \\ 
        \midrule
        8 & 0.26 & 74.1 & 0.73 \\
        16 & 0.53 & 87.0 & 0.89 \\
        32 & 1.06 & 90.7 & 0.92 \\ 
        \bottomrule
    \end{tabular}
    }
    % \vspace{-3mm}
\end{table}

%% file: sec/7_conclusion.tex
\section{Discussion}

The CGV dataset and its accompanying framework are fundamentally motivated by the potential of artificial intelligence to democratize and enhance pediatric healthcare~\cite{cao2025workshop,liang2019evaluation}. A primary objective is to automate clinical gait evaluation, ultimately scaling these capabilities to unconstrained, ``in-the-wild'' environments. Currently, traditional 2D observational tools, such as the Edinburgh Visual Gait Score (EVGS), rely heavily on expert visual inspection. This reliance introduces inherent subjectivity and inter-rater variability, limiting reproducibility across different evaluators and institutions. Furthermore, human-administered video assessments are also highly sensitive to environmental confounders. 

To overcome these limitations, we establish the first comprehensive visual benchmark for children's gait analysis. We evaluate the efficacy of contemporary foundation models and introduce a specialized, multi-stage pipeline, ChildGait-Video, integrating instance segmentation, anatomical pose estimation, and spatiotemporal modeling. Empirical results demonstrate that our framework achieves an agreement with expert annotations ranging from 70\% to 93\% in all items for typical and atypical distinction, with an average 0.83 F1 score. Crucially, unlike marker-based 3D motion capture systems that demand expensive, specialized infrastructure, our approach operates on standard RGB camera setups. This drastically lowers both technical and economic barriers, facilitating scalable deployment across heterogeneous and resource-constrained clinical settings.

While the current study establishes robust model-level validation, future efforts will extend this framework to open-scene data collected directly from home, rehabilitation centers, and community environments. This expansion will improve algorithmic generalization across a broader spectrum of complex pathological gait patterns and unconstrained acquisition scenarios. Deploying such technology in the wild enables reproducible longitudinal tracking and cross-site standardization, which are essential for large-scale outcome evaluation and data-driven rehabilitation research.  Ultimately, the integration of robust computer vision methodologies into routine pediatric care heralds a paradigm shift: moving from subjective, observer-dependent evaluations toward highly scalable, objective, and computationally reproducible clinical workflows~\cite{cao2023vitasd,cao2023commentary}.

\section{Conclusion}
In this work, we introduce fine-grained pediatric gait understanding from standard RGB videos as a new computer vision problem and present CGV, a large-scale multi-view dataset with synchronized anonymized pose, segmentation, and clinically grounded EVGS-derived annotations. Through systematic benchmarking, we show that contemporary zero-shot MLLMs and fine-tuned VLMs remain unreliable for phase-sensitive clinical gait scoring. To address these challenges, we propose ChildGait-Video, an end-to-end adaptation paradigm. This design substantially improves per-item EVGS classification. We expect CGV and the proposed framework to serve as a rigorous benchmark for pediatric gait research and to facilitate scalable. 

%% file: sec/X_suppl.tex
\appendix
\clearpage
\section*{Appendix}\label{sec:appendix}

\section{CGV Details}
\label{sec:details}

\begin{figure}[!h]
    \centering
    \begin{subfigure}[b]{0.45\textwidth}
        \centering
        \includegraphics[height=4cm]{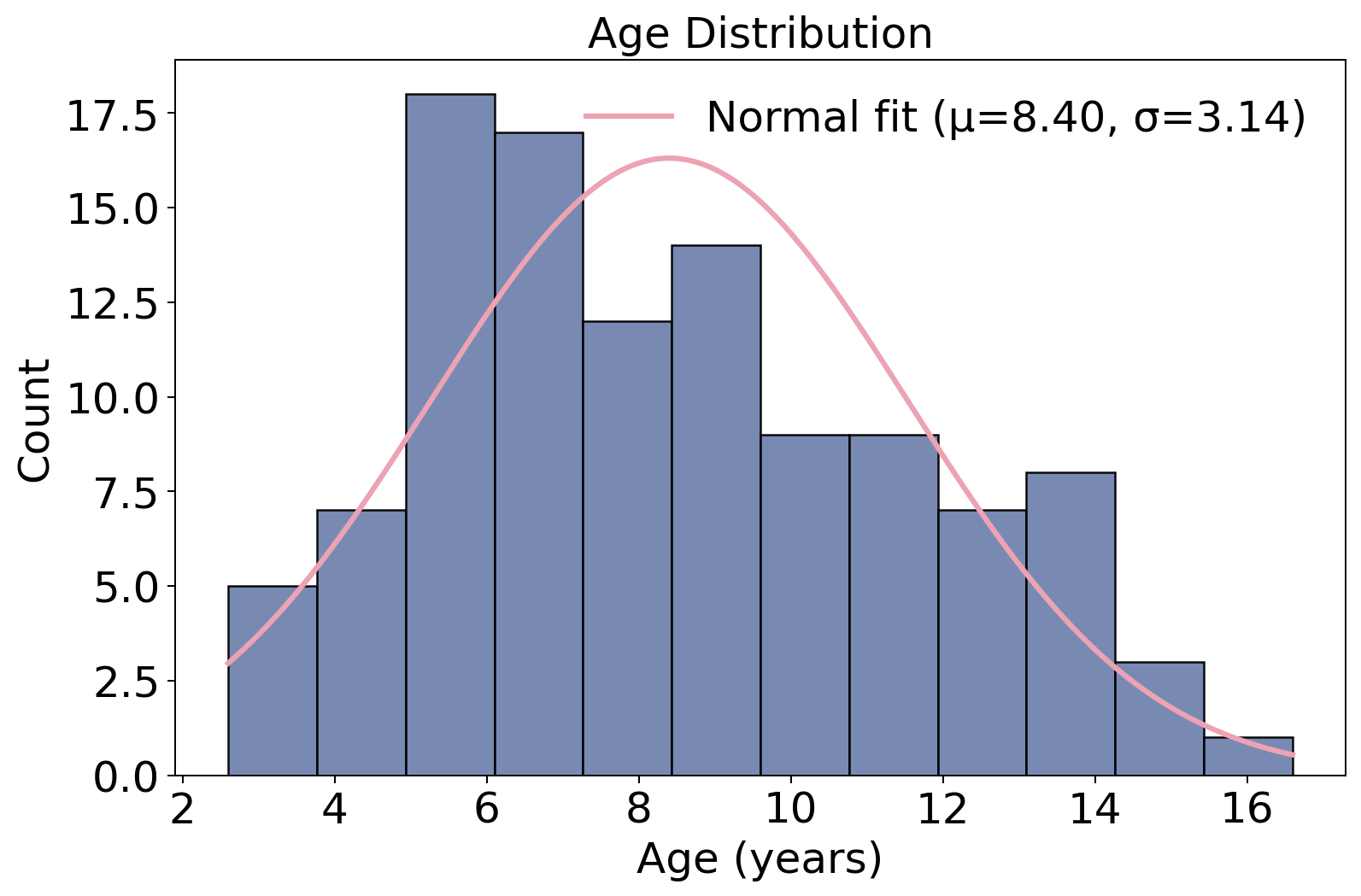} 
        \caption{\textbf{Age Distribution.}}
        \label{fig:age_distribution}
    \end{subfigure}
    \hfill
    \begin{subfigure}[b]{0.45\textwidth}
        \centering
        \includegraphics[height=4cm]{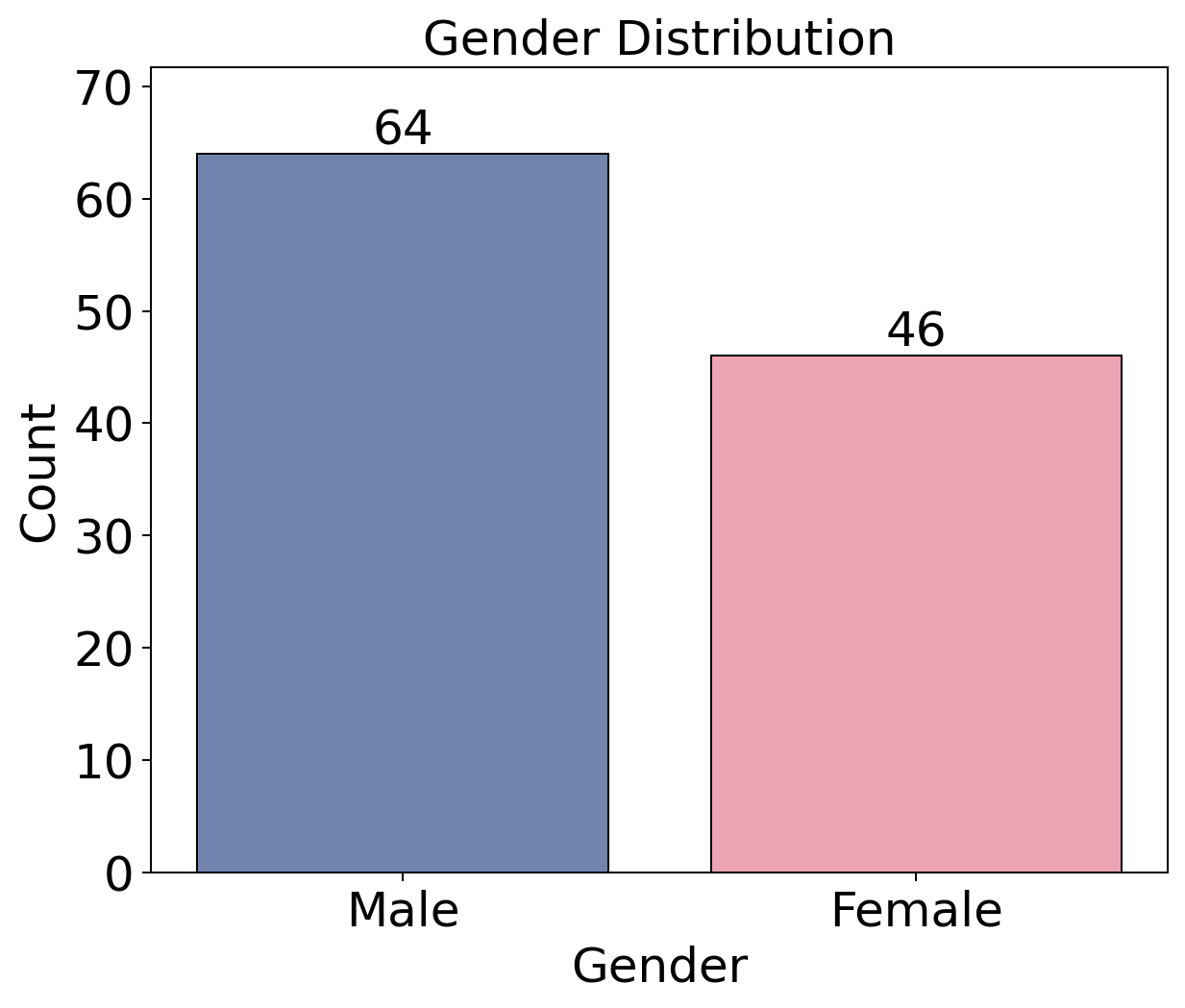} 
        \caption{\textbf{Gender Distribution.}}
        \label{fig:gender_distribution}
    \end{subfigure}
    \caption{\textbf{Demographic Statistics of the ChildGait Video Dataset.} \textit{left:} The age distribution of the subjects with a fitted normal distribution curve, highlighting a primary focus on the pediatric population. \textit{right:} The gender distribution across the entire dataset. Overall, the dataset maintains a nearly balanced demographic profile, providing the necessary diversity for training or evaluation in visual gait analysis.}
    \label{fig:demographics}
\end{figure}

In addition to the overall dataset scale, annotation procedures, and label distributions discussed in~\Cref{sec:dataset}, we provide further details regarding the demographic composition of the participants in the Children Gait Video (CGV) dataset. As illustrated in~\Cref{fig:demographics}, the dataset consists of  110 pediatric patients, aiming to represent the target clinical population as fully as possible. 

\noindent \textbf{Age Distribution.}~\Cref{fig:age_distribution} visualizes the age profile of the patients. The histogram is overlaid with a fitted normal distribution curve to illustrate the central tendency of the patient demographics. The participants span a critical developmental window from 2.6 to 16.6 years old, with a mean age of $\mu=8.40, \sigma=3.14$ years, which is highly clinically relevant and highlights a primary focus on the pediatric population. To explore the relationship between age and performance, we split the test set into younger children ($\le 8$ years) and older children ($\geq 8$ years) subsets, achieving average accuracies of 82.4\% and 85.2\%, respectively, proving that gait analysis is more difficult for younger children.

\noindent \textbf{Gender Distribution.}~\Cref{fig:gender_distribution} details the gender composition of the dataset. The patients consist of 58.2\% male and 41.8\% female subjects. This nearly balanced gender distribution helps to prevent the model from learning shortcuts for specific genders, ensuring sufficient diversity and fairness for training and evaluating models for visual gait analysis.

\section{EVGS Scoring Criteria}
\label{sec:evgs}

\input{tabs/scale}

The Edinburgh Visual Gait Score (EVGS)~\cite{read2003edinburgh} is a clinical tool developed to visually assess gait deviations in ambulatory children using coronal and sagittal video recordings. It evaluates 17 observational scoring items for each limb that are graded on a three-point ordinal scale. \Cref{tab:scale} details this specific ordinal scale used for the evaluation process. \Cref{tab:scores} provides the comprehensive clinical criteria, explanations, and score mappings for the gait scoring items.

\input{tabs/scores}

\input{figs/prompts}

\section{Prompt Design}
\label{sec:prompts}

As stated in~\Cref{sec:benchmarking} and~\Cref{sec:vlm_gait}, we design a structured prompt to instruct the MLLMs to perform visual gait analysis. The prompt architecture is composed of three core components: System Persona, Task Instructions, and Output Formatting.

\noindent \textbf{System Persona.} We set the model's role as an expert pediatrician specializing in observational gait analysis. This initialization ensures that the model leverages its pre-trained domain knowledge of anatomical priors, gait cycle phases, and kinematic deviations.

\noindent \textbf{Task Instructions.} We explicitly define the evaluation workflow, asking the model to systematically process both coronal and sagittal video recordings to assess the 17 scoring items for each limb. To ground the model's reasoning and ensure standardized evaluation, we embed the exact EVGS scoring criteria shown in~\Cref{sec:evgs} into the context. The model evaluates each item on a three-point ordinal scale, where 0 indicates a normal condition, 1 represents a moderate deviation, and 2 signifies a marked deviation, as stated in~\Cref{sec:evgs}. We enforce that the model uniformly outputs 1 for both 1 and 2 to align with our requirements mentioned in~\Cref{sec:dataset}. \looseness=-1

\noindent \textbf{Output Formatting.} The model is asked to return its assessment in a strict JSON format for easy evaluation and inspection. \looseness=-1

Overall, the designed prompts are shown in~\Cref{fig:prompts}, where Scoring Item, Explanation, and Scoring Criteria all originate from~\Cref{tab:scores}.

\section{Evaluation of Skeleton Baselines}
\label{sec:skeleton}

\input{tabs/skeleton_results}

To evaluate the performance of skeleton baselines in the field of children's gait analysis, we evaluate strong baselines SkeletonGait++~\cite{fan2024skeletongait} and ScoNet~\cite{zhou2024gait, zhou2025pose}.

\noindent \textbf{Experimental Results.} As shown in~\Cref{tab:skeleton_results}, although general skeleton-based baselines have demonstrated strong capabilities in standard pose and gait recognition tasks, they still struggle to perform precise clinical scoring consistently across all joints, yielding sub-optimal average accuracies that range from 58\% to 65\% for the left limb and 60\% to 64\% for the right limb. Notably, ScoNet pre-trained on Scoliosis1K achieves the best results, peaking at average accuracies of 68\% and 70\% for the left and right limbs, respectively. We attribute the limited performance of the baselines to the inherent design objectives of traditional skeleton-based networks, which focus on extracting global structural representations for macro-level classification, thereby neglecting the fine-grained, localized kinematic anomalies essential for clinical assessment, especially at distal joints (\eg, FRT, POB), where accuracies drop sharply. \looseness=-1

\begin{table}[h]
    \centering
    \caption{\textbf{Module Ablation Study Results.} We evaluate the individual contributions of each proposed module, including Token-Level Kinematic Prompting (TKP) and Mask-Guided Patch Pruning (MPP), against various masking baselines.}
    \label{tab:module}
    \setlength{\tabcolsep}{8pt} 
     \resizebox{0.8\linewidth}{!}{%
     \begin{tabular}{lcccc}
        \toprule
        \textbf{Variant} & \textbf{L-AVG (\%)} & \textbf{R-AVG (\%)} & \textbf{L-F1} & \textbf{R-F1} \\
        \midrule
        VideoMAE v2 (Base)               & 69 & 72 & 0.70 & 0.71 \\
        VideoMAE v2 + TKP                & 72 & 74 & 0.72 & 0.73 \\
        VideoMAE v2 + MPP                & 78 & 79 & 0.77 & 0.78 \\
        VideoMAE v2 + Random Mask        & 68 & 70 & 0.69 & 0.69 \\
        VideoMAE v2 + Bounding Box Mask  & 75 & 77 & 0.74 & 0.76 \\
        \midrule
        \textbf{ChildGait-Video (Ours)}  & \textbf{84} & \textbf{84} & \textbf{0.83} & \textbf{0.83} \\
        \bottomrule
    \end{tabular}}
\end{table}

\section{Module Ablation Study}
\label{sec:module}

To comprehensively evaluate the individual contributions of our proposed modules and justify our design choices, we conduct a detailed module ablation study. As shown in~\Cref{tab:module}, Token-Level Kinematic Prompting (TKP) helps to gain a performance improvement, raising the average accuracy (L-AVG/R-AVG) from 69\%/72\% to \textbf{72\%/74\%}. This demonstrates that TKP effectively provides necessary anatomical priors derived from expert annotations, guiding the network to focus on clinically relevant joint dynamics rather than generic spatial features. We also validate the Mask-Guided Patch Pruning (MPP) module designed for noise mitigation. We compare our MPP with two alternative strategies: a standard \textit{bounding box} mask and a \textit{random} mask. While the \textit{bounding box} mask brings a moderate gain by removing some background (achieving 75\%/77\% for L/R-AVG), our MPP outperforms it by \textbf{3\%} and \textbf{2\%} on the left and right limbs, respectively, proving its superior capability in adaptively filtering background noise while preserving structural integrity. Conversely, employing the \textit{random} mask strategy degrades the overall performance, causing a drop of \textbf{1\%} to \textbf{2\%} compared to the baseline (down to 68\%/70\%). This decline is expected, as random masking inevitably obscures crucial kinematic joints or fails to adequately eliminate background noise, thereby destroying the details essential for accurate clinical assessment.

%% file: tabs/scale.tex
\begin{table*}[!h]
    \centering
    \caption{\textbf{EVGS Three-point Scale.} Each scoring item is assessed and categorized into one of three scales: Normal, Moderate deviation, or Marked deviation.}
    % \vspace{-3mm}
    \label{tab:scale}
    \resizebox{0.8\linewidth}{!}{%
    \begin{tabular}{cl}
        \toprule
        \multicolumn{1}{c}{\textbf{Ordinal Scale}} & \multicolumn{1}{c}{\textbf{Measurement}} \\
        \midrule
        0 & Normal (within +/- 1.5 standard deviations (SD) of normal mean) \\
        1 & Moderate deviation (between 1.5 and 4.5 SD of normal mean) \\
        2 & Marked deviation (greater than 4.5 SD of normal mean) \\
        \bottomrule
    \end{tabular}}
    % \vspace{-3mm}
\end{table*}

%% file: tabs/scores.tex
\begin{center}
    \footnotesize
    \begin{longtable}{
        @{}
        >{\centering\arraybackslash}m{0.05\textwidth}
        >{\centering\arraybackslash}m{0.15\textwidth}
        >{\noindent\justifying\arraybackslash}m{0.42\textwidth}
        >{\raggedright\arraybackslash}m{0.28\textwidth}
        @{}
    }  
        \caption{\textbf{EVGS Items and Scores.} We list all EVGS scoring items, including explanations and scoring details.}
        % \vspace{-5mm}
        \label{tab:scores} \\
        \\
        \toprule
        \multicolumn{1}{c}{\textbf{No.}} & \multicolumn{1}{c}{\textbf{Scoring Items}} & \multicolumn{1}{c}{\textbf{Explanation}} & \multicolumn{1}{c}{\textbf{Score}} \\ 
        \midrule
		\endfirsthead

        \toprule
        \multicolumn{1}{c}{\textbf{No.}} & \multicolumn{1}{c}{\textbf{Scoring Items}} & \multicolumn{1}{c}{\textbf{Explanation}} & \multicolumn{1}{c}{\textbf{Score}} \\ 
        \midrule
		\endhead

        \midrule
        \endfoot
        
        \bottomrule
        \endlastfoot

        1 & Initial Contact in Stance & The heel normally contacts first. The toe describes that portion of the foot distal to the metatarsophalangeal joints. Simultaneous contact with the heel and toe comprises flatfoot contact. & \begin{itemize}[leftmargin=*, nosep]
            \item Heel contact: 0
            \item Flatfoot contact: 1
            \item Toe contact: 2
        \end{itemize} \\ 

        \\

        2 & Heel Lift in Stance & If there is no heel contact during stance, there can be no heel lift (i.e., `No heel contact'). Heel lift normally occurs between the opposite foot level and the opposite foot contact (`Normal'). `Early' heel lift indicates that heel lift precedes the opposite foot being level with the stance foot. `Delayed' heel lift is present if heel lift occurs with or after opposite foot contact. `No forefoot contact' describes the rare occasion of a calcaneus foot when the forefoot does not contact during stance. & \begin{itemize}[leftmargin=*, nosep]
            \item No forefoot contact: 2
            \item Delayed: 1
            \item Normal: 0
            \item Early: 1
            \item No heel contact: 2
        \end{itemize} \\

        \\ 

        3 & Max Ankle Dorsiflexion in Stance & There is normal forward progression of the tibia over the planted hind-foot from slight plantar flexion at initial contact to dorsiflexion at terminal stance. Describe the maximum angle of dorsiflexion between the hind foot and the shaft of the tibia during stance. In pathological gait, lack of heel contact may be caused by either excessive plantar flexion of the foot or excessive knee flexion. The tibial hind-foot angle is therefore analyzed irrespective of the position of the foot on the floor. & \begin{itemize}[leftmargin=*, nosep]
            \item Excessive dorsiflexion (>40° df): 2
            \item Increased dorsiflexion (26°- 40° df): 1
            \item Normal dorsiflexion (5°- 25° df): 0
            \item Reduced dorsiflexion (10° pl - 4° df)
            \item Marked plantar flexion (>10° pl): 2
        \end{itemize} \\

        \\ 

        4 & Hind-foot Varus/Valgus in Stance & In the coronal plane, the normal hind-foot is in neutral or very slight valgus. & \begin{itemize}[leftmargin=*, nosep]
            \item Severe valgus (>15° valgus): 2
            \item Mod valgus (6°- 15° valgus): 1
            \item Neutral/slight valgus (0°- 5° valgus): 0
            \item Mild varus (1°- 10° varus): 1
            \item Severe varus (>10° varus): 2
        \end{itemize} \\ 

        \\ 

        5 & Foot Rotation in Stance & The normal foot is slightly externally rotated relative to the Knee Progression Angle (KPA, i.e., the direction in which the knee points during gait). & \begin{itemize}[leftmargin=*, nosep]
            \item Marked ext. >KPA (by >40°): 2
            \item Mod ext. >KPA (by 21°- 40°): 1,
            \item Slightly more ext. than KPA (by 0°- 20° extension): 0
            \item Mod int. >KPA (by 1°- 25°): 1
            \item Marked int. >KPA (by >25°): 2
        \end{itemize} \\

        \\ 

        6 & Foot Clearance in Swing & The whole foot, including the toe, should clear the foot and not make contact during the swing phase. 'None' should be recorded if there is continuous contact between some part of the foot and the floor throughout the swing phase. `Reduced' indicates that there is a shortened but definite period of clearance during some part of the swing phase between the whole foot and the floor. 'Full' or normal clearance is when the foot does not touch at all in swing; however, normal clearance is a very small amount. `High steps' describes excessive lifting of the foot from the floor. When there is reduced clearance followed by high stepping, circle both, giving a score of 2 for this combination of features. & \begin{itemize}[leftmargin=*, nosep]
            \item High Steps: 1
            \item Full: 0
            \item Reduced: 1
        \end{itemize} \\

        \\ 

        7 & Max Ankle Dorsiflexion in Swing & The ankle is normally approximately neutral in swing, but very slight plantar flexion (5°) is acceptable. & \begin{itemize}[leftmargin=*, nosep]
            \item Excessive dorsiflexion (>30° df): 2
            \item Increased dorsiflexion (16°- 30° df): 1
            \item Normal dorsiflexion (15° df - 5° pl): 0
            \item Mod plantar flexion (6°- 20° pl): 1
            \item Marked plantar flexion (>20° pl): 2
        \end{itemize} \\

        \\

        8 & Knee Progression Angle in Mid-Stance & The knee normally points forward during gait. Record the position in which the knee appears to point during most of the stance phase. When either internal or external rotation is present, but the whole knee cap is visible, score 1. When rotation is present to such an extent that the knee cap is partially out of view (external or internal, part of the cap visible), score 2. & \begin{itemize}[leftmargin=*, nosep]
            \item External, part of the knee cap visible: 2
            \item External, all of the knee cap visible: 1
            \item Neutral, knee cap midline: 0
            \item Internal, all of the knee cap visible: 1
            \item Internal, part of the knee cap visible: 2
        \end{itemize} \\

        \\

        9 & Peak Knee Extension in Stance & The knee approaches full extension in terminal stance. In pathological gait, the knee may remain more flexed throughout stance. Alternatively, hypertension can occur as femoral progression proceeds over an arrested tibia. & \begin{itemize}[leftmargin=*, nosep]
            \item Severe flexion (>25°): 2
            \item Mod flexion (16°- 25°): 1
            \item Normal (0°- 15° flexion): 0
            \item Mod hyperextension (1°- 10°): 1
            \item Severe hyperextension (<10°): 2
        \end{itemize} \\

        \\

        10 & Knee Position in Terminal Swing & The knee is normally in slight flexion immediately before heel strike. & \begin{itemize}[leftmargin=*, nosep]
            \item Severe flexion (>30°): 2
            \item Mod flexion (16°- 30°): 1
            \item Normal (5°- 15° flexion): 0
            \item Mod overextension (4° flexion - 10° extension): 1
            \item Severe hyperextension (>10° extension): 2
        \end{itemize} \\

        \\

        11 & Peak Knee Flexion in Swing & The normal range is 50° to 70°. & \begin{itemize}[leftmargin=*, nosep]
            \item Severely increased (>85° flexion): 2
            \item Mod increased (71°- 85° flexion): 1
            \item Normal (50°- 70° flexion): 0
            \item Mod reduced (35°- 49° flexion): 1
            \item Severely reduced (<35° flexion): 2
        \end{itemize} \\

        \\ 

        12 & Peak Hip Extension in Stance & The hip normally extends in stance to between neutral and 20° of extension. & \begin{itemize}[leftmargin=*, nosep]
            \item Severe flexion (>15° flexion): 2
            \item Mod flexion (1°- 15° flexion): 1
            \item Normal (0°- 20° extension): 0
            \item Mod hyperextension (21°- 35° extension): 1
            \item Marked hyperextension (>35° extension): 2
        \end{itemize} \\ 

        \\ 

        13 & Peak Hip Flexion during Swing & Normal flexion is between 25° and 45°. & \begin{itemize}[leftmargin=*, nosep]
            \item Marked increased flexion (>60° flexion): 2
            \item Increased flexion (46°- 60° flexion): 1
            \item Normal flexion (25°- 45° flexion): 0
            \item Reduced flexion (10°- 24° flexion): 1
            \item Severely reduced (<10° flexion): 2
        \end{itemize} \\

        \\

        14 & Pelvic Obliquity at Mid-Stance & The pelvis normally drops slightly on the opposite side during loading, becoming level by terminal stance. Estimate the position in mid stance. `Up' and `down' refer to the position of the ASIS on the stance side, relative to the opposite side ASIS. & \begin{itemize}[leftmargin=*, nosep]
            \item Marked down (>10°): 2
            \item Mod down (1°- 10°): 1
            \item Normal obliquity (0°- 5° up): 0
            \item Mod up (6°- 15°): 1
            \item Marked up (>15°): 2
        \end{itemize} \\

        \\

        15 & Pelvic Rotation at Mid-Stance & In mid stance, the pelvis should be at approximately neutral rotation, between 5° backward rotation (retraction) of the stance leg, and 10° forward rotation (protraction). & \begin{itemize}[leftmargin=*, nosep]
            \item Marked retraction (>15°): 2
            \item Mod retraction (6°- 15°): 1
            \item Normal (5° retraction - 10° protraction): 0
            \item Mod protraction (11°- 20°): 1
            \item Marked protraction (>20°): 2   
        \end{itemize} \\

        \\

        16 & Peak Sagittal Trunk Position in Stance & The trunk is erect during the stance and swing phases. & \begin{itemize}[leftmargin=*, nosep]
            \item Marked forward lean (>15° forward): 2
            \item Mod forward lean (between 6° and 15° forward): 1
            \item Normal upright (vertical to 5° forward or backward): 0
            \item Mod backward lean (>5° backward): 1
        \end{itemize} \\

        \\

        17 & Maximum Trunk Lateral Shift & Normally, the trunk displaces laterally approximately 25 mm during stance, towards the stance leg. `Excessive' thoracic shift laterally or lateral flexion should be considered when recording observations. `Reduced' describes those cases in which the trunk remains leaning over the swinging leg. & \begin{itemize}[leftmargin=*, nosep]
            \item Marked: 2
            \item Mod: 1
            \item Normal: 0
            \item Reduced: 1
        \end{itemize} \\

        \\
        
    \end{longtable}
\end{center}

%% file: figs/prompts.tex
\begin{figure*}[!h]
    \centering
    \resizebox{0.99\textwidth}{!}{
        \begin{planbox}{Designed Prompts for Visual Gait Analysis}
            \small

            You are a highly skilled and experienced \textbf{pediatrician} specializing in gait analysis. Your expertise lies in evaluating and diagnosing gait abnormalities in children through advanced computational methods. You possess deep knowledge of biomechanics, pediatric orthopedics, and motion analysis technologies. \newline
            \textbf{Your Aim:} analyze children's gait patterns in the video provided, and give a score for each parameter of the Edinburgh Visual Gait Score (EVGS). \newline
            
            \textbf{Scoring Item:} \texttt{<Item>} \newline
            \textbf{Explanation:} \texttt{<Explanation>} \newline
            \textbf{Scoring Criteria:} \texttt{<Scores>} \newline

            Based on the above criteria, please analyze the gait pattern of the child in the video and provide a detailed EVGS score report. For each scoring item, give a score along with a brief explanation of your assessment. \newline

            Output the results in the following JSON format: \newline
            \texttt{\{} \newline
            \hspace*{0.5cm} \texttt{"items": [} \newline
            \hspace*{1.0cm} \texttt{\{} \newline
            \hspace*{1.5cm} \texttt{"Item": str,} \newline
            \hspace*{1.5cm} \texttt{"Explanation": str,} \newline
            \hspace*{1.5cm} \texttt{"Score": int} \newline
            \hspace*{1.0cm} \texttt{\},} \newline
            \hspace*{1.0cm} \texttt{...} \newline
            \hspace*{0.5cm} \texttt{]} \newline
            \texttt{\}}

            \hrulefill

            \textbf{Input Data}: \newline
            \hspace*{0.5cm} \textbf{Video Input:} \texttt{<Video Frames>} \newline
            \hspace*{0.5cm} \textbf{Prompts:} \texttt{<Prompts>}

            \textbf{Output Data:} \newline
            \hspace*{0.5cm} \textbf{JSON File:} \texttt{<JSON File>}
            
        \end{planbox}
    }
    \caption{\textbf{Designed Prompts for Visual Gait Analysis.} The model takes both the video frames and designed prompts as input to generate an explanation and a score for each scoring item.}
    \label{fig:prompts}
\end{figure*}

%% file: tabs/skeleton_results.tex
\begin{table*}[!t]
    \centering
    \caption{\textbf{Quantitative Evaluation of Skeleton Baselines.} All reported values are \textbf{percentages (\%)}. We report 17 scoring items for the \textbf{Left (L-)} limb (top) and \textbf{Right (R-)} limb (bottom). The detailed definition of each item is shown in~\Cref{tab:dataset_item}. L/R-AVG indicates the average accuracy over all scoring items of the left/right limb. L/R-F1 indicates the average F1-score across all items of the left/right limb.}
    \label{tab:skeleton_results}
    % \vspace{-3mm}
    \setlength{\tabcolsep}{3.3pt}
    
    % --- Table: Left Limb ---
    \resizebox{0.99\linewidth}{!}{%
        \begin{tabular}{@{} ll *{19}{c} @{}}
            \toprule
            \textbf{Method} & \textbf{Pretrain} & \mrot{L-IC} & \mrot{L-HL} & \mrot{L-SAD} & \mrot{L-WAD} & \mrot{L-HVV} & \mrot{L-FRT} & \mrot{L-FCL} & \mrot{L-KPA} & \mrot{L-KEX} & \mrot{L-KPS} & \mrot{L-KFX} & \mrot{L-HEX} & \mrot{L-HFX} & \mrot{L-POB} & \mrot{L-PRT} & \mrot{L-TSG} & \mrot{L-TLT} & \mrot{L-AVG} & \mrot{L-F1} \\
            \midrule
            SkeletonGait++~\cite{fan2024skeletongait} & Gait3D~\cite{zheng2022gait} & 68 & 70 & 63 & 67 & 53 & 48 & 50 & 54 & 60 & 56 & 62 & 64 & 61 & 43 & 46 & 54 & 67 & 58 & 0.44 \\
            SkeletonGait++~\cite{fan2024skeletongait} & SUSTech1K~\cite{shen2023lidargait} & 75 & 77 & 70 & 73 & 59 & 54 & 57 & 61 & 68 & 63 & 70 & 71 & 68 & 52 & 54 & 61 & 72 & 65 & 0.61 \\
            SkeletonGait++~\cite{fan2024skeletongait} & GREW~\cite{zhu2021gait} & 70 & 73 & 66 & 69 & 56 & 51 & 53 & 58 & 64 & 59 & 65 & 67 & 64 & 48 & 50 & 56 & 68 & 61 & 0.54 \\
            SkeletonGait++~\cite{fan2024skeletongait} & CCPG~\cite{li2023depth} & 68 & 70 & 63 & 67 & 53 & 48 & 49 & 55 & 60 & 56 & 62 & 64 & 61 & 44 & 45 & 54 & 67 & 58 & 0.47 \\
            ScoNett~\cite{zhou2024gait, zhou2025pose} & Scoliosis1Kt~\cite{zhou2024gait, zhou2025pose} & 77 & 80 & 73 & 76 & 63 & 58 & 60 & 64 & 71 & 66 & 73 & 74 & 71 & 54 & 57 & 64 & 75 & 68 & 0.67 \\
            \bottomrule
        \end{tabular}
    }
    
    \vspace{2mm}
        
    % --- Table: Right Limb + Avg Acc ---
    \resizebox{0.99\linewidth}{!}{%
        \begin{tabular}{@{} ll *{19}{c} r @{}}
            \toprule
            \textbf{Method} & \textbf{Pretrain} & \mrot{R-IC} & \mrot{R-HL} & \mrot{R-SAD} & \mrot{R-WAD} & \mrot{R-HVV} & \mrot{R-FRT} & \mrot{R-FCL} & \mrot{R-KPA} & \mrot{R-KEX} & \mrot{R-KPS} & \mrot{R-KFX} & \mrot{R-HEX} & \mrot{R-HFX} & \mrot{R-POB} & \mrot{R-PRT} & \mrot{R-TSG} & \mrot{R-TLT} & \mrot{R-AVG} & \mrot{R-F1} \\
            \midrule
            SkeletonGait++~\cite{fan2024skeletongait} & Gait3D~\cite{zheng2022gait} & 71 & 75 & 67 & 69 & 57 & 52 & 55 & 58 & 65 & 60 & 66 & 67 & 64 & 49 & 50 & 58 & 71 & 62 & 0.60 \\
            SkeletonGait++~\cite{fan2024skeletongait} & SUSTech1K~\cite{shen2023lidargait} & 73 & 77 & 69 & 71 & 59 & 54 & 57 & 59 & 67 & 62 & 68 & 69 & 66 & 51 & 53 & 60 & 73 & 64 & 0.53 \\
            SkeletonGait++~\cite{fan2024skeletongait} & GREW~\cite{zhu2021gait} & 73 & 77 & 68 & 72 & 58 & 54 & 57 & 60 & 66 & 63 & 68 & 69 & 66 & 51 & 52 & 61 & 73 & 64 & 0.55 \\
            SkeletonGait++~\cite{fan2024skeletongait} & CCPG~\cite{li2023depth} & 69 & 72 & 65 & 67 & 55 & 51 & 53 & 56 & 62 & 58 & 63 & 65 & 63 & 47 & 48 & 56 & 70 & 60 & 0.46 \\
            ScoNett~\cite{zhou2024gait, zhou2025pose} & Scoliosis1Kt~\cite{zhou2024gait, zhou2025pose} & 79 & 82 & 75 & 77 & 65 & 60 & 63 & 67 & 75 & 70 & 76 & 77 & 74 & 59 & 61 & 69 & 81 & 70 & 0.69 \\
            \bottomrule
        \end{tabular}
    }
    % \vspace{-3mm}
\end{table*}